\documentclass[letterpaper]{article} 
\usepackage{aaai2026}  
\usepackage{times}  
\usepackage{helvet}  
\usepackage{courier}  
\usepackage[hyphens]{url}  
\usepackage{graphicx} 
\usepackage{natbib}  
\usepackage{caption} 
\usepackage{algorithm}
\usepackage{algorithmic}

\usepackage{amsmath,amsfonts}
\usepackage{eucal}
\usepackage{colortbl}
\usepackage{booktabs}
\usepackage{multirow}
\usepackage{makecell}
\usepackage{pifont}

\usepackage{newfloat}
\usepackage{listings}
\DeclareCaptionStyle{ruled}{labelfont=normalfont,labelsep=colon,strut=off} 
\floatstyle{ruled}
\newfloat{listing}{tb}{lst}{}
\floatname{listing}{Listing}
\title{LWGANet: Addressing Spatial and Channel Redundancy in Remote Sensing Visual Tasks with Light-Weight Grouped Attention}
\author{
	Wei Lu\textsuperscript{\rm 1},
	Xue Yang\textsuperscript{\rm 2},
	Si-Bao Chen\textsuperscript{\rm 1*}
}
\affiliations{
	\textsuperscript{\rm 1}MOE Key Lab of ICSP, IMIS Lab of Anhui, Anhui Provincial Key Lab of Multimodal Cognitive Computation, Zenmorn-AHU AI Joint Lab, School of Computer Science and Technology, Anhui University, Hefei 230601, China\\
	\textsuperscript{\rm 2}School of Automation and Intelligent Sensing, Shanghai Jiao Tong University, Shanghai 200240, China\\
	E23101009@stu.ahu.edu.cn, yangxue-2019-sjtu@sjtu.edu.cn, sbchen@ahu.edu.cn
}

\usepackage{bibentry}

\begin{document}

\maketitle
\begin{abstract}	\label{abstract}
	Light-weight neural networks for remote sensing (RS) visual analysis must overcome two inherent redundancies: spatial redundancy from vast, homogeneous backgrounds, and channel redundancy, where extreme scale variations render a single feature space inefficient. Existing models, often designed for natural images, fail to address this dual challenge in RS scenarios. To bridge this gap, we propose LWGANet, a light-weight backbone engineered for RS-specific properties. LWGANet introduces two core innovations: a Top-K Global Feature Interaction (TGFI) module that mitigates spatial redundancy by focusing computation on salient regions, and a Light-Weight Grouped Attention (LWGA) module that resolves channel redundancy by partitioning channels into specialized, scale-specific pathways. By synergistically resolving these core inefficiencies, LWGANet achieves a superior trade-off between feature representation quality and computational cost. Extensive experiments on twelve diverse datasets across four major RS tasks---scene classification, oriented object detection, semantic segmentation, and change detection---demonstrate that LWGANet consistently outperforms state-of-the-art light-weight backbones in both accuracy and efficiency. Our work establishes a new, robust baseline for efficient visual analysis in RS images.
\end{abstract}
\vspace{-2mm}
\begin{links}
	\link{Code}{https://github.com/AeroVILab-AHU/LWGANet}
\end{links}

\section{Introduction} \label{sec_Introduction}
\begin{figure}[t]	\centering
	\includegraphics[width=1\linewidth]{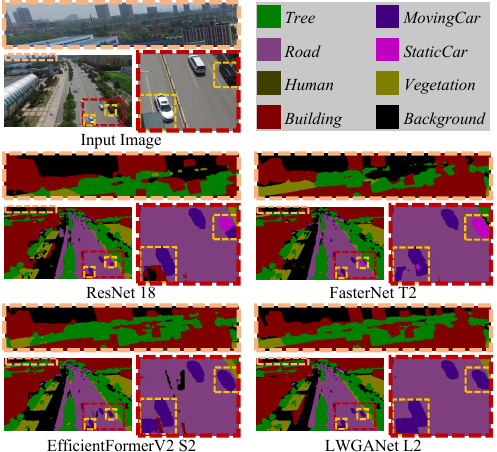}
	\caption{Visual comparison on the UAVid testing set~\cite{lyu2020uavid} with UnetFormer~\cite{unetformer} as the decoder. FasterNet~\cite{chen2023run}, a convolution-based model, excels at capturing building details but struggles with the global context needed for segmenting moving cars. Conversely, EfficientFormer V2~\cite{Eff_Formerv2}, leveraging global attention, effectively segments cars but fails to preserve fine-grained building structures. Our LWGANet achieves a superior balance by jointly modeling local details and long-range dependencies.}	\label{visual_UAVid} \vspace{-5mm}
\end{figure}

The efficiency of deep learning models~\cite{Lu_2025_ICCV,LU2026431,11021615,11058953,10990319,11202372,3MOS} for remote sensing (RS) image analysis is inherently constrained by two types of data redundancy. First, \textbf{spatial redundancy} arises from the sparse distribution of salient foreground objects within vast, homogeneous backgrounds such as roads, farmland, or oceans. Naive dense computation over the entire image leads to disproportionate resource allocation toward these background regions, which contribute minimal semantic value.

Second, a more subtle but critical issue is \textbf{channel redundancy}, stemming from the extreme scale variations in RS imagery. A single, unified feature representation struggles to capture both fine-grained textures and broad spatial contexts efficiently. For instance, channels specialized for small objects like vehicles are underutilized when processing large structures like runways, and vice-versa. This forces a compromise where a significant portion of the feature space becomes irrelevant for any given scale, leading to computational waste and representational inefficiency.

While existing lightweight backbone networks---such as MobileNetV2~\cite{mobilenetv2}---have been widely adopted in RS applications (e.g., scene classification~\cite{zhang2019lightweight,LGRIN}, object detection~\cite{LO-Det,LiteSalNet}, semantic segmentation~\cite{RSR-Net,LENet}, and change detection~\cite{you2024robust}), these architectures are primarily designed for natural image benchmarks like ImageNet~\cite{imagenet}. Their efficiency typically derives from simplifications based on homogeneous grouping---applying identical operators like depthwise separable convolutions across all channel partitions. This uniform approach, while reducing parameters, fundamentally fails to address channel redundancy in RS data. It forces a single operational logic onto a diverse, multi-scale feature space, leading to computational waste and representational compromises. As a result, they often fail to reconcile the competing demands of local detail preservation and long-range context modeling, especially under the multi-scale and cluttered conditions of RS images.

This architectural misalignment leads to a performance trade-off. For instance, convolution-based models such as FasterNet~\cite{chen2023run} exhibit strong local representation but lack the receptive field to capture global dependencies, resulting in poor recognition of diffuse objects. Conversely, transformer-style models like EfficientFormer V2~\cite{Eff_Formerv2} possess enhanced global modeling capabilities but often suppress the high-frequency spatial information critical for detecting small or detailed objects. These limitations motivate the need for a lightweight backbone explicitly designed to mitigate both spatial and channel redundancies in a synergistic manner.

In this work, we propose \textbf{LWGANet}, a novel light-weight backbone architected to resolve this dual redundancy problem in RS images. It is built upon two key principles: \\
(1) \textbf{To address spatial redundancy}, we propose the \textbf{Top-K Global Feature Interaction (TGFI)} module. This component selectively samples a sparse set of informative spatial positions and performs global context aggregation on this reduced token set. The result is a resolution-independent mechanism for long-range dependency modeling with reduced computational cost. \\
(2) \textbf{To alleviate channel redundancy}, we design the \textbf{Light-Weight Grouped Attention (LWGA)} module. It partitions channels into heterogeneous groups, routing each through a specialized pathway optimized for a distinct feature scale---ranging from fine-grained edges to high-level semantics. This structure enables simultaneous multi-scale representation while minimizing channel waste.

Through the integration of TGFI and LWGA, LWGANet achieves a balanced capacity for both fine spatial detail extraction and large-scale semantic modeling. As shown in Figure \ref{visual_UAVid}, the network effectively parses diverse RS scenes containing extreme scale variation and complex background interference. In summary, our contributions are as follows:
\begin{itemize}
	\item We identify spatial and channel redundancy as key bottlenecks for efficient network design in RS images, providing an architectural strategy to tackle both jointly.
	\item We propose the LWGA module, a novel grouped attention architecture that resolves channel redundancy by decoupling features into specialized, multi-scale pathways.
	\item We introduce the TGFI module, a sparse interaction mechanism that mitigates spatial redundancy by efficiently modeling global context on a reduced set of salient features.
	\item We present LWGANet, a new light-weight backbone built upon these principles, and validate its state-of-the-art performance and versatility through extensive experiments on 12 datasets across four distinct RS tasks.
\end{itemize}

\section{Related Work}		\label{Related_Work}
The pursuit of efficient deep learning models has led to a variety of light-weight architectures. Early CNN-based efforts, such as MobileNetV2~\cite{mobilenetv2}, utilized depthwise separable convolutions to reduce computational cost. More recent techniques include model pruning~\cite{zheng2022model}, knowledge distillation~\cite{hinton2015distilling}, reparameterization~\cite{ding2021repvgg}, and Neural Architecture Search (NAS)~\cite{MnasNet}.

Concurrently, Vision Transformers (ViTs)~\cite{vit} and their hierarchical variants such as PVT \cite{wang2021pyramid} and Swin Transformer~\cite{liu2021swin} have introduced a new paradigm. Light-weight ViTs, such as MobileViT~\cite{mobilevit} and EfficientFormerV2~\cite{Eff_Formerv2}, aim to merge transformer capabilities with mobile efficiency. Despite these advancements, most generic light-weight backbones are primarily optimized on natural image datasets like ImageNet~\cite{imagenet}, which lack the extreme scale variations and high redundancy characteristic of RS imagery.

Although some light-weight CNNs and ViTs have been applied to RS tasks, their performance often plateaus due to their general-purpose nature. While these general-purpose architectures have been adapted for RS tasks, they were not engineered to exploit RS-specific properties. Moreover, they rarely exploit the significant feature redundancy present in RS data. Consequently, a critical need remains for light-weight backbones specifically engineered for RS imagery---architectures that can manage channel redundancy and spatial redundancy. Our proposed LWGANet fills this gap by systematically addressing these challenges through decoupled multi-scale feature representation and sparse global context modeling.


\section{Approach} \label{LWGANet_all}
This section presents the overall architecture of LWGANet and its core components.

\subsection{Overview of LWGANet} \label{sec_architecture}
LWGANet adopts a hierarchical architecture with four stages, progressively reducing the spatial resolution by factors of 4, 8, 16, and 32. This multi-scale design is fundamental for handling the diverse object scales prevalent in RS imagery. To accommodate different computational budgets, we propose three variants---L0, L1, and L2---distinguished by their stem layer channel counts (32, 64, and 96), providing a clear trade-off between model capacity and efficiency.

The architecture begins with a stem layer, implemented as a stride-4 convolution, to quickly reduce spatial dimensions while expanding channel capacity. Each stage then comprises a sequence of LWGA blocks, with block counts $N_1, N_2, N_3, N_4$ set to [1, 2, 4, 2] for L0 and L1, and [1, 4, 4, 2] for L2. For downsampling between stages, we employ the DRFD module~\cite{lu2023robust}, chosen for its proven ability to preserve fine details. The multi-level features produced by each stage can be readily fed into task-specific decoders for various downstream applications.

Within each stage, an LWGA block processes the input feature map $\mathbf{X}$ first through the LWGA module, producing an enhanced feature map $\mathbf{Y}$. This is followed by a Channel Multilayer Perceptron (CMLP) that refines $\mathbf{Y}$ using sequential 1$\times$1 convolutions for channel expansion and restoration. The block is then completed with a residual connection, Batch Normalization (BN), and dropout:
\begin{equation}
	\begin{aligned}
		\text{CMLP}(\mathbf{Y}) &= \text{Conv}(\text{Act}(\text{BN}(\text{Conv}(\mathbf{Y})))), \\
		\text{out} &= \mathbf{X} + \text{BN}(\text{drop}(\text{CMLP}(\mathbf{Y}))),
	\end{aligned}
	\label{eq:cmlp}
\end{equation}
where the dropout rate is 0.0 for L0, 0.1 for L1 and L2.

\begin{figure}[t]
	\centering
	\includegraphics[width=0.99\linewidth]{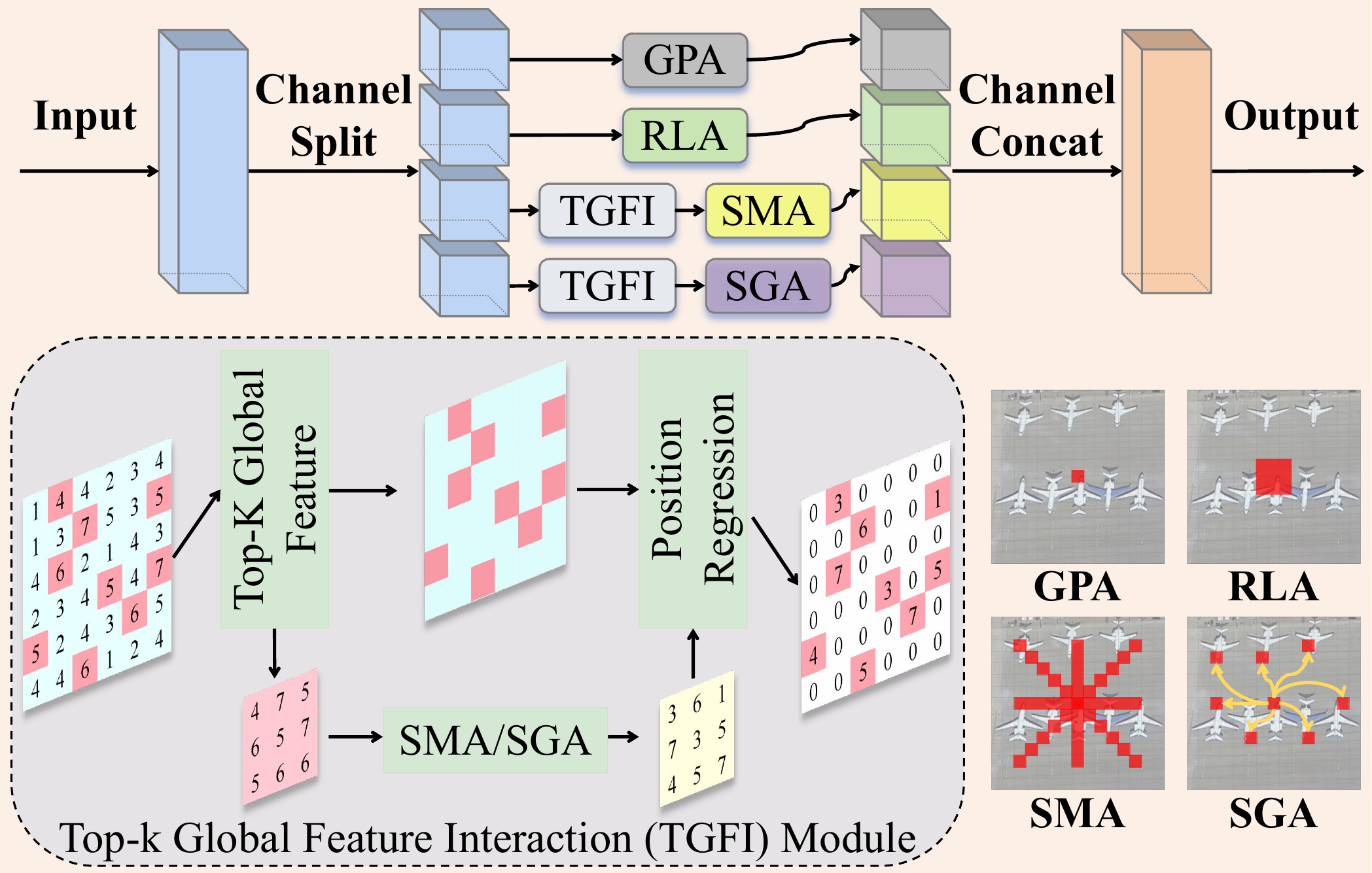}
	\caption{Illustration of the LWGA module and its submodules. GPA, RLA, SMA and SGA denote gate point attention, regular local attention, sparse medium-range attention and sparse global attention, respectively.}	
	\label{fig_lwga_module}  \vspace{-5mm}
\end{figure}

\subsection{LWGA Module} \label{sec_lwga_module}
The LWGA module is the core engine of LWGANet, engineered to resolve channel redundancy through a novel heterogeneous grouping strategy. Traditional lightweight designs often employ homogeneous grouping (e.g., grouped convolutions or multi-head attention), where identical operations are applied to all channel partitions. This uniform processing is inefficient for the extreme scale variations in RS imagery. In contrast, LWGA decouples the feature space by partitioning channels into four specialized, non-overlapping pathways $\{\mathbf{X_1}, \mathbf{X_2}, \mathbf{X_3}, \mathbf{X_4}\}$, where each segment $\mathbf{X_{i}} \in \mathbb{R}^{H \times W \times C/4}$ is routed through a distinct operator optimized for a specific feature scale.

This channel partitioning allows us to maximize representational efficiency. Each channel group is allocated to a specialized submodule optimized for a distinct feature scale, ensuring that the allocated channels are used with maximum relevance and minimal waste. The four pathways are designed to capture:\\
\textbf{Point-level Details:} $\mathbf{X_1}$ is processed by Gate Point Attention (GPA) to enhance fine-grained features.\\
\textbf{RLA for Local Patterns:}
$\mathbf{X_2}$ is handled by Regular Local Attention (RLA), which uses standard convolutions. This approach leverages their strong inductive bias for local texture and pattern recognition.\\
\textbf{Medium-range Structures:} $\mathbf{X_3}$ is fed into Sparse Medium-range Attention (SMA) to capture contextual information for irregularly shaped objects.\\
\textbf{Global Context:} $\mathbf{X_4}$ is processed by Sparse Global Attention (SGA) to model long-range dependencies for overall scene understanding.

By assigning a specialized, computationally efficient operator to each scale-specific task, the LWGA module avoids the compromises inherent in one-size-fits-all approaches. The outputs $\{\mathbf{R_1}, \mathbf{R_2}, \mathbf{R_3}, \mathbf{R_4}\}$ are then concatenated, fusing the multi-scale representations into a comprehensive feature map $\mathbf{Y} \in \mathbb{R}^{H \times W \times C}$. Figure \ref{fig_lwga_module} provides a detailed view of the structure of each submodule.

\subsubsection{Top-K Global Feature Interaction (TGFI) Module.}
Before detailing the SMA and SGA modules, we first introduce TGFI, a key submodule designed to mitigate \textbf{spatial redundancy} for efficient long-range dependency modeling. TGFI is motivated by a core inefficiency in RS imagery: naive global attention mechanisms expend immense computational effort on vast, uninformative background regions.
To address this inefficiency, TGFI implements a sparse interaction strategy that focuses computation exclusively on the most informative features, thus overcoming the limitations imposed by spatial redundancy. As shown in Figure \ref{fig_lwga_module}:\\
\textbf{(1) Sparse Feature Sampling:} It first divides the input features into non-overlapping regions and selects the single most salient feature token (e.g., with the highest activation value) from each region. The spatial coordinates $\mathcal{P}_{loc}$ of these selected features are preserved. This effectively creates a compact yet representative summary of the features.\\
\textbf{(2) Subspace Interaction:} Next, interactions (e.g., via convolutions or attention) are computed only among this reduced set of sampled features. This establishes global relationships in an efficient subspace, drastically reducing the complexity compared to processing the full feature map.\\
\textbf{(3) Feature Restoration:} Finally, the enhanced representations are restored to their original spatial locations using the preserved coordinates $\mathcal{P}_{loc}$, while non-sampled locations are typically filled via interpolation or identity mapping.

By leveraging the sparse nature of RS data, TGFI serves as an intelligent sampling and interaction mechanism. It not only significantly reduces computational costs but also minimizes interference from irrelevant background noise, enabling efficient and robust global context integration.

\subsubsection{GPA Module.} \label{sec_gpa}

The GPA module is designed to highlight fine-grained details critical for detecting small objects or intricate textures in RS imagery. It begins with a 1$\times$1 convolution to expand the input feature $\mathbf{X_{1}}$ from $C/4$ to $C$ channels, enabling richer feature representations. The expanded features undergo batch normalization and activation. Subsequently, a second 1$\times$1 convolution restores the channel dimension to $C/4$, yielding $\mathbf{X_{1}'}$. A sigmoid function is applied to $\mathbf{X_{1}'}$ to generate an attention map $\mathcal{A}_{1}$, which weights the importance of each feature. The output $\mathbf{R_1}$ is computed as $\mathbf{R_{1}} = \mathbf{X_{1}} + \mathcal{A}_{1} \cdot \mathbf{X_1}$.

\subsubsection{RLA Module.}
The RLA module is designed to efficiently capture local spatial features, leveraging the strong inductive bias of convolutions for local pattern recognition. The input feature $\mathbf{X_{2}}$ undergoes a 3$\times$3 convolution (with parameters \textit{I}=C/4, \textit{O}=C/4, \textit{K}=3, \textit{S}=1, \textit{P}=1), preserving spatial dimensions. Batch normalization and a non-linear activation follow to produce the output feature $\mathbf{R_{2}}$. This provides a robust foundation for modeling local dependencies.

\subsubsection{SMA Module.}
The SMA module is engineered to capture medium-range contextual information, which is essential for objects with irregular shapes or those spanning beyond the immediate local neighborhood. It operates as follows: the TGFI module first reduces the input feature $\mathbf{X_{3}}$ to $\mathbf{X_{3}'} \in \mathbb{R}^{\frac{H}{3} \times \frac{W}{3} \times \frac{C}{4}}$, expanding the receptive field while preserving positional coordinates $\mathcal{P}_{loc}$. The resulting $\mathbf{X_{3}'}$ is processed to generate an attention map $\mathcal{A}_{3}'$ that integrates contextual information, computed as follows:
\begin{equation}
	\begin{aligned}
\hspace{-3mm}	\mathcal{A}_{ij} &\hspace{-1mm}=\hspace{-3mm} \sum_{n=0}^{n=\frac{L-1}{2}}\hspace{-3mm} \alpha_{(i \pm n)j} \hspace{-1mm}\cdot\hspace{-1mm} x_{(i \pm n)j}
		\hspace{-1mm}+\hspace{-3mm} \sum_{n=0}^{n=\frac{L-1}{2}}\hspace{-3mm} \alpha_{(i \pm n)(j \pm n)} \hspace{-1mm}\cdot\hspace{-1mm} x_{(i \pm n)(j \pm n)}\\
		&\hspace{-1mm} +\hspace{-3mm} \sum_{n=0}^{n=\frac{L-1}{2}}\hspace{-3mm} \alpha_{i(j \pm n)} \hspace{-1mm}\cdot\hspace{-1mm} x_{i(j \pm n)}
		\hspace{-1mm}+\hspace{-3mm} \sum_{n=0}^{n=\frac{L-1}{2}}\hspace{-3mm} \alpha_{(i \mp n)(j \pm n)} \hspace{-1mm}\cdot\hspace{-1mm} x_{(i \mp n)(j \pm n)},
	\end{aligned}
\end{equation}
where $x_{ij}$ represents the feature at position $(i, j)$, and $\alpha_{ij}$ are learnable coefficients. The window size $L$ is set to 11. The attention map $\mathcal{A}_{3}'$ is then interpolated back to the original feature map dimensions using the preserved coordinates $\mathcal{P}_{loc}$, yielding $\mathcal{A}_{3}$ ($\mathcal{A}_{3}' \in \mathbb{R}^{\frac{H}{3} \times \frac{W}{3} \times \frac{C}{4}} \stackrel{\mathcal{P}_{loc}}{\longrightarrow}\mathcal{A}_3 \in \mathbb{R}^{H \times W \times \frac{C}{4}}$). $\mathbf{R_{3}}$ is computed as: $\mathbf{R_{3}} = \mathcal{A}_{3} \cdot \mathbf{X_{3}}$.

\subsubsection{SGA Module.}
The SGA module is responsible for capturing long-range dependencies and global context. To balance representation with computational feasibility across stages, it adopts a \textbf{dynamic, stage-aware strategy} that adapts to the decreasing spatial resolution of the feature maps.

\textbf{For Stages 1 and 2,} where feature maps are large, a standard self-attention mechanism is computationally prohibitive due to its quadratic complexity with respect to the number of tokens. Even on the sparsely sampled features from TGFI, this cost remains a bottleneck. To circumvent this, we employ a highly efficient proxy for global attention: a combination of a 5$\times$5 grouped convolution and a 7$\times$7 dilated convolution (dilation=3). This convolutional approach approximates long-range interactions with a complexity that is linear to the number of sampled tokens, offering a powerful yet light-weight solution for capturing expansive context in high-resolution stages. The TGFI module first samples foreground features, reducing the feature map to $\mathbf{X_{4}'} \in \mathbb{R}^{\frac{H}{2} \times \frac{W}{2} \times \frac{C}{4}}$ while preserving positional coordinates $\mathcal{P}_{loc}$. The convolutional attention map $\mathcal{A}_{412}$ is applied to this sparse feature set, and the resulting feature $\mathbf{R_{4}'} = \mathcal{A}_{412} \cdot \mathbf{X_{4}'}$ is then restored to the original resolution using $\mathcal{P}_{loc}$, i.e., $\mathbf{R_{4}'} \in \mathbb{R}^{\frac{H}{2} \times \frac{W}{2} \times \frac{C}{4}} \stackrel{\mathcal{P}_{loc}}{\longrightarrow}\mathbf{R_{4}''} \in \mathbb{R}^{H \times W \times \frac{C}{4}}$, and the final output is computed as $\mathbf{R_{4}} = \text{BN}(\mathbf{R_{4}''} + \mathbf{X_{4}})$.

\textbf{For Stage 3,} as the feature map size is considerably reduced, we can afford a more powerful interaction mechanism. The TGFI module again reduces the feature map to $\mathbf{X_{4}'} \in \mathbb{R}^{\frac{H}{2} \times \frac{W}{2} \times \frac{C}{4}}$. A standard global self-attention mechanism~\cite{vit} with 4 heads, denoted $\mathcal{A}_{43}$, is then applied to $\mathbf{X_{4}'}$. The output $\mathbf{R_{4}'} = \mathcal{A}_{43} \cdot \mathbf{X_{4}'}$ is interpolated back to the full resolution via $\mathcal{P}_{loc}$, and combined with the input: $\mathbf{R_{4}} = \text{BN}(\mathbf{R_{4}''} + \mathbf{X_{4}})$.

\textbf{For Stage 4,} the feature map is at its smallest spatial dimension, containing highly condensed semantic information. At this final stage, computational cost is no longer the primary constraint. We therefore apply the full-power standard global self-attention directly to the entire (dense) feature map $\mathbf{X_{4}}$, without sparse sampling. This maximalist approach ensures that the model can perform a comprehensive, all-to-all comparison of semantic concepts across the entire scene, which is vital for final classification and high-level understanding. The output is computed as $\mathbf{R_{4}} = \text{BN}(\mathcal{A}_{44} \cdot \mathbf{X_{4}} + \mathbf{X_{4}})$, where $\mathcal{A}_{44}$ denotes the standard global self-attention.

This progressive strategy, moving from highly efficient convolutional approximations to sparse attention and finally to dense global attention, allows LWGANet to model global context with a sophistication that scales gracefully with the network's depth and feature map size.

Finally, the features from all four pathways, $\{\mathbf{R_{1}}, \mathbf{R_{2}}, \mathbf{R_{3}}, \mathbf{R_{4}}\}$, are concatenated along the channel dimension. This fusion step integrates the specialized, multi-scale information into a single, comprehensive feature map $\mathbf{Y} \in \mathbb{R}^{H \times W \times C}$, ready for subsequent processing.

\section{Experiments}	\label{Exp}
In this section, we conduct a comprehensive evaluation of LWGANet on 12 public datasets across four key tasks: \textbf{scene classification} on UCM~\cite{yang2010bag}, AID~\cite{xia2017aid}, and NWPU-RESISC45~\cite{NWPU}; \textbf{oriented object detection} on DOTA-v1.0/1.5~\cite{dota} (online testing) and DIOR-R~\cite{AOPG}; \textbf{semantic segmentation} on UAVid~\cite{lyu2020uavid} and LoveDA~\cite{loveda} (both online testing); and \textbf{change detection} on LEVIR-CD~\cite{chen2020spatial}, WHU-CD~\cite{ji2018fully}, CDD-CD~\cite{lebedev2018change}, and SYSU-CD~\cite{shi2021deeply}.

\subsection{Datasets and Experimental Setup}
Across all experiments, data preprocessing and training protocols aligned with established methods~\cite{lu2024decouplenet,cai2024pkinet,unetformer,A2Net,CLAFA}, maintaining standard dataset splits and applying the best validation weights for testing. We use a backbone pre-trained on ImageNet-1K~\cite{imagenet}. Detailed dataset statistics and implementation specifics are provided in the Appendix for full reproducibility, accessible at: \url{https://github.com/AeroVILab-AHU/LWGANet/blob/main/figures/LWGANet_sup.pdf}.

\begin{table}[t] \centering	\scriptsize	
	\renewcommand{\arraystretch}{0.96}
\setlength{\tabcolsep}{0.5mm}{
\begin{tabular}{ccc|ccc|ccc} \toprule
\multirow{2}{*}{Method}
&\multirow{2}{*}{\makecell[c]{Params. \\ (M) $\downarrow$} }
& \multirow{2}{*}{\makecell[c]{FLOPs \\ (G) $\downarrow$}}
& \multicolumn{3}{c}{Top-1 Accuracy  (\%) $\uparrow$} \vline
& \multicolumn{3}{c}{Speed (FPS) $\uparrow$ } \\ \cline{4-9}
&&&NWPU&AID&UCM& GPU&CPU&ARM\\ \midrule										
MobileNet V2 1.0$\times$& 2.28  & 0.319 & 95.06 & 93.65	& 97.14	& 11301	& 49.11 & 785.4	\\ 
FasterNet T0			& 2.68  & 0.338 & 93.30 & 92.85	& 94.75	& 18276	& 106.4	& 839.7	\\
StarNet	S1				& 2.68	& 0.431	& 94.30	& 91.05	& 93.10	& 6045	& 71.70	& 459.8	\\
EdgeViT XXS				& 3.79  & 0.546 & 94.75 & 93.10	& 95.24	& 4153	& 46.36	& 259.9	\\
EfficientformerV2 S0	& 3.36  & 0.396 & 94.52 & 93.80	& 97.14	& 1299	& 54.00	& 272.0	\\
EdgeNeXt XXS			& 1.17  & 0.197 & 92.35 & 88.10	& 88.10	& 8521	& 100.8	& -		\\
$^{\star}$MobileViT XXS	& 1.03  & 0.333 & 94.37 & 93.30	& 95.00	& 4811	& 31.87	& 453.6	\\
GhostNet V2 0.6$\times$	& 2.16  & 0.077 & 94.65 & 92.70	& 94.29	& 9802	& 69.96	& 867.1	\\
\textbf{LWGANet-L0}&1.72&0.186&\textbf{95.49}&\textbf{94.60}&\textbf{98.57}&13234&80.00&687.8 \\
\midrule	
MobileNet V2 2.0$\times$& 8.81  & 1.17 	& 95.35 & 93.85	& 97.86	& 5567	& 17.03	& 372.3	\\
FasterNet T1			& 6.37  & 0.855 & 93.73 & 93.20	& 94.52	& 11876	& 51.42	& 650.1	\\
StarNet	S3				& 5.5	& 0.767	& 93.32	& 91.40	& 93.33	& 4438	& 47.86	& 336.8	\\
EdgeViT XS				& 6.40  & 1.12 	& 94.89 & 93.75	& 94.52	& 3310	& 29.95	& 205.8	\\
PVT V2 B0				& 3.42  & 0.533 & 94.35 & 93.10	& 96.43	& 3843	& 40.26	& 243.4	\\
EfficientformerV2 S1	& 5.87  & 0.661 & 94.97 & 93.95	& 96.90	& 1211	& 36.96	& 204.5	\\
EdgeNeXt XS				& 2.15  & 0.408 & 92.79 & 90.45	& 88.10	& 5455	& 56.42	& -		\\
$^{\star}$MobileViT XS 	& 2.02  & 0.900 & 94.90 & \textbf{95.20}	& 96.43	& 3300	& 12.99	& 306.3	\\
GhostNet V2 1.0$\times$	& 4.93  & 0.181 & 95.08 & 93.80	& 94.76	& 6596	& 42.02 & 591.7	\\
\textbf{LWGANet-L1}&5.90&0.709&\textbf{95.70}&94.85&\textbf{98.81}&6418&34.08&375.8	\\
\midrule	
MobileNet V2 2.5$\times$& 13.7 	& 1.80 	& 95.48 & 94.45	& 97.86	& 3796	& 12.90	& 282.4	\\
FasterNet T2			& 13.8 	& 1.91 	& 95.11 & 93.60	& 94.29	& 6852	& 26.40	& 669.8	\\
StarNet	S4				& 7.23	& 1.07	& 93.08	& 89.75	& 90.71	& 3093	& 34.20	& 235.1	\\
EdgeViT S 				& 12.7 	& 1.90 	& 95.05 & 93.35	& 95.95	& 2318	& 19.31	& 141.3	\\
PVT V2 B1 				& 13.5 	& 2.04 	& 94.62 & 93.45	& 95.71 & 2369	& 15.96	& 145.3	\\
EfficientformerV2 S2	& 12.3 	& 1.26 	& 95.14 & 94.20	& 97.38	& 642	& 24.74	& 123.8	\\
EdgeNeXt S 				& 5.30 	& 0.960 & 93.54 & 91.90	& 92.62	& 3844	& 30.00	& -		\\
$^{\star}$MobileViT S	& 5.03 	& 1.75 	& 95.19 & 95.25	& 97.14	& 2681	& 10.20	& 152.7	\\
GhostNet V2 2.0$\times$	& 16.7 	& 0.632 & 95.44 & 94.30	& 95.95	& 3476	& 21.32	& 303.2	\\
\textbf{LWGANet-L2}&13.0&1.87&\textbf{96.17}&\textbf{95.45}&\textbf{98.57}&3308&16.18&274.3 \\
\bottomrule
\end{tabular}	}\caption{Experimental results on NWPU, AID, and UCM classification datasets with a training image size of 224$\times$224. The symbol `$\star$' indicates a training image size of 256$\times$256. FPS were acquired by the RTX 3090 (GPU), Intel i9-11900K (CPU), and NVIDIA AGX-XAVIER (ARM) with batch sizes of $256$, $16$, and $32$, respectively.} \label{table_cla}
\end{table}

\subsection{Evaluation on Scene Classification} \label{Cla}
We begin by evaluating LWGANet on scene classification, a task that directly tests the model's multi-scale representation capability. Table \ref{table_cla} presents a comprehensive comparison against SOTA light-weight models, including MobileNet V2~\cite{mobilenetv2}, FasterNet~\cite{chen2023run}, StarNet~\cite{StarNet}, PVT V2~\cite{pvtv2}, EdgeViT~\cite{pan2022edgevits}, EfficientformerV2~\cite{Eff_Formerv2}, EdgeNeXt~\cite{edgenext}, MobileViT~\cite{mobilevit}, and GhostNet V2~\cite{ghostnetv2}.

Our LWGANet variants consistently achieve strong performance across all three datasets. For instance, the light-weight LWGANet-L0 demonstrates a compelling balance of accuracy and efficiency, achieving 95.49\% Top-1 accuracy on the challenging NWPU dataset. This result surpassed light-weight models like StarNet S1 (+1.19\%) while using significantly fewer parameters and FLOPs.

This trend of high accuracy relative to computational cost continues with our larger variants, L1 and L2, which set new benchmarks for light-weight models on these datasets. The consistent high performance, from the texture-rich scenes in UCM to the complex layouts in NWPU, suggests that our LWGA module's feature decoupling strategy effectively captures the required spectrum of features. The model excels because it does not compromise local detail for global context, or vice versa, validating our core design principle.

\subsubsection{Analysis of Practical Efficiency.}
To assess practical deployment performance, we benchmarked inference speeds (FPS) across GPU, CPU, and ARM platforms, as shown in Table \ref{table_cla}. The results highlight LWGANet's excellent practical efficiency. For instance, our LWGANet-L0 achieved a high throughput of 13,234 FPS on GPU and a strong 80.0 FPS on CPU, outperforming most competing hybrid and Transformer-based models like EfficientformerV2 and EdgeViT by a significant margin. This demonstrates that our design is not only theoretically efficient but also translates to tangible speed in practice.
FasterNet, a pure CNN-based model, achieves higher FPS due to its architecture being composed almost entirely of highly optimized standard convolutions. Despite this, LWGANet offers a superior trade-off: it delivered substantially higher accuracy than FasterNet (+2.19\% on NWPU) while maintaining highly competitive, real-world inference speeds, making it a more balanced and powerful solution for practical RS applications.

\begin{table}[t]	\centering	\scriptsize	
\renewcommand{\arraystretch}{0.96}
\setlength{\tabcolsep}{1mm}{
\begin{tabular}{c|cc|ccc|c} \toprule	
\multirow{2}{*}{Method}
&\multirow{2}{*}{\makecell[c]{$\#$P \\ (M) $\downarrow$}}
&\multirow{2}{*}{\makecell[c]{FLOPs\\(G) $\downarrow$}}
&\multicolumn{4}{c}{mAP ($\%$)  $\uparrow$}	\\  \cline{4-7}
&&							& DOTA1.0	& DOTA1.5	& DIOR-R	& Mean\\  \hline	
ResNet-50 	& 41.1	& 211.4	& 75.87		& 66.88		& 64.30		& 69.02\\
FasterNet-T2& 30.0	& 160.3	& 76.17		& 71.07		& 63.66		& 70.30\\
ARC-R50		& 74.4	& 212.0	& 77.35		& 68.31		& 65.51		& 70.39\\
LSKNet-S 	& 31.0	& 161.0	& 77.49 	& 70.26		& 65.90		& 71.22\\
EfficientFormerV2-S2&29.2&145.1&76.70	& 72.19		& 65.00		& 71.30\\
DecoupleNet-D2&\textbf{23.3}	& \textbf{142.4}	& 78.04		& 71.15 	& 67.08		& 72.09\\
PKINet-S 	& 30.8	& 184.6	& 78.39		& 71.47		& 67.03		& 72.30\\
\textbf{LWGANet-L2}&29.2 &159.1 &\textbf{79.02}&\textbf{72.91}&\textbf{68.53}&\textbf{73.49}\\ \bottomrule
\end{tabular}	}	
\caption{Experimental results on the DOTA 1.0, DOTA 1.5, and DIOR-R test sets with single-scale training and testing.}
\label{tab_Det_DOTA10_15_DIORR}
\end{table}	

\begin{table}[t]
	\centering
	\scriptsize
	\renewcommand{\arraystretch}{0.96}
	\setlength{\tabcolsep}{2mm}{
		\begin{tabular}{c|cc|cc}	\toprule
			\multirow{2}{*}{Method}
			&\multicolumn{2}{c}{Backbone only} \vline
			&\multirow{2}{*}{\makecell[c]{Speed\\(FPS) $\uparrow$}}
			&\multirow{2}{*}{\makecell[c]{mAP\\($\%$) $\uparrow$}}  \\	
			&$\#$P (M) $\downarrow$&FLOPs (G) $\downarrow$ && \\ \hline
EfficientFormerV2-S2& \textbf{12.0}	& \textbf{26.8}	 &15.7&76.70													\\
			ARC-R50 & 56.5 & 86.6 & 11.8 & 77.35 \\
			LSKNet-S & 14.4 & 54.4 & \textbf{22.5} & 77.49 \\
			PKINet-S & 13.7 & 70.2 & 5.4 & 78.39 \\
\textbf{LWGANet-L2} & \textbf{12.0} & 38.8 & 19.4 & \textbf{79.02} \\	\bottomrule
	\end{tabular}}\caption{Speed and accuracy comparison of different backbones on the DOTA 1.0 test set~\cite{dota}.} \label{tab_Det_DOTA10_compar}	
\end{table}

\begin{table*}[ht]
	\begin{minipage}{0.48\linewidth}  \centering	\scriptsize	
		\renewcommand{\arraystretch}{1}
		\setlength{\tabcolsep}{1.6mm}{
			\begin{tabular}{c|cccc} \toprule
				Method &$\#$P $\downarrow$ & FLOPs $\downarrow$ &FPS $\uparrow$ &mIoU $\uparrow$\\ \hline
EfficientFormerV2-S2~\cite{Eff_Formerv2}	&12.7& \textbf{35.8}& 54.1 &65.2	\\
				FasterNet-T2~\cite{chen2023run}	&13.3&49.1& \textbf{198.8}	&65.7	\\
				ResNet18~\cite{he2016deep}	& \textbf{11.7}	& 46.9			& 115.6	& 67.8 \\
\textbf{LWGANet-L2}&12.6&50.3	& 67.3	& \textbf{69.1}	\\ \bottomrule
		\end{tabular}	}
	
	\end{minipage} 	\hspace{3mm}
	\begin{minipage}{0.48\linewidth} \centering \scriptsize
		\renewcommand{\arraystretch}{0.96}
		\setlength{\tabcolsep}{4mm}{
			\begin{tabular}{c|cc} \toprule
				Method & $\#$P (M) $\downarrow$ & mIoU $\uparrow$ \\
				\hline
				FactSeg~\cite{factseg} & 33.44 & 50.0 \\
				FarSeg~\cite{FarSeg} & 31.37 & 50.1 \\
				LoveNAS~\cite{lovenas} & 30.49 & 52.3 \\
				UnetFormer~\cite{unetformer} & \textbf{11.73} & 52.4 \\
				RSSFormer~\cite{rssformer} & 30.82 & 52.4 \\
\textbf{LWGANet-L2} & 12.58 & \textbf{53.6} \\
				\bottomrule
		\end{tabular} }	
	\end{minipage}\caption{Segmentation experimental results on the UAVid test set~\cite{lyu2020uavid} (left) and LoveDA~\cite{loveda} (right) compared to the SOTA methods. $\#$P, FLOPs, and speeds were tested with a 1,024$\times$1,024 input size on the RTX 3090 GPU.} \label{tab_seg_UAVid_loveda}
\end{table*}

\begin{table*}[t]	\centering	\scriptsize
\renewcommand{\arraystretch}{0.96}
\setlength{\tabcolsep}{1mm}{
\begin{tabular}{ccc|ccc|ccc|ccc|ccc|ccc} \toprule
\multirow{2}{*}{Method}
&\multirow{2}{*}{\makecell[c]{$\#$P \\ (M) $\downarrow$}}
&\multirow{2}{*}{\makecell[c]{FLOPs\\(B) $\downarrow$}}
&\multicolumn{3}{c}{LEVIR-CD}\vline	&\multicolumn{3}{c}{WHU-CD}\vline
&\multicolumn{3}{c}{CDD-CD}\vline	&\multicolumn{3}{c}{SYSU-CD}\vline
&\multicolumn{3}{c}{Mean}  \\ \cline{4-18}
&&&IoU $\uparrow$&F1 $\uparrow$&Pre $\uparrow$&IoU $\uparrow$&F1 $\uparrow$&Pre $\uparrow$
&IoU $\uparrow$&F1 $\uparrow$&Pre $\uparrow$&IoU $\uparrow$&F1 $\uparrow$&Pre $\uparrow$
&IoU $\uparrow$&F1 $\uparrow$&Pre $\uparrow$ \\ \hline	
BIT \cite{BIT} &3.50&10.6
&81.30&89.69&92.33&78.24&87.79&85.07&70.22&82.50&89.03&60.48&75.38&80.08&72.56&83.84&86.63\\
DMINet \cite{DMINet}&6.24&14.6
&82.99&90.75&92.52&79.68&88.69&93.84&86.91&92.99&93.03&51.56&68.04&64.47&75.29&85.12&85.97\\
RFANet \cite{you2024robust} &\textbf{2.86}&3.16
&84.32&91.49&92.32&87.86&93.54&95.46&87.04&93.07&93.84&69.08&81.71&82.06&82.08&89.95&90.92\\

A2Net~\cite{A2Net}&3.78&3.05
&84.21&91.43&92.08&88.98&94.17&\textbf{96.68}&\textbf{87.42}&\textbf{93.29}&93.65&70.83&82.93&\textbf{86.45}&82.86&90.46&92.22\\
\textbf{A2Net-LWGANet-L0}&2.91&\textbf{2.76}
&\textbf{84.94}&\textbf{91.86}&\textbf{92.81}&\textbf{90.24}&\textbf{94.87}&96.11&87.24&93.18&\textbf{94.30}&\textbf{71.54}&\textbf{83.41}&86.16&\textbf{83.49}&\textbf{90.83}&\textbf{92.35}\\ \midrule	

IFNet \cite{IFNet}&35.7&82.3
&80.36&89.11&91.64&73.16&84.50&87.92&85.68&92.29&92.03&64.42&78.36&78.49&75.91&86.07&87.52\\
ChangeFormer \cite{changerformer}&41.0&203
&81.59&89.86&91.12&72.92&84.34&89.08&85.40&92.13&93.39&64.20&78.20&79.70&76.03&86.13&88.32\\
ICIFNet \cite{icif}&23.8&25.4
&81.75&89.96&91.32&79.24&88.32&92.98&85.73&92.31&93.41&40.65&57.80&71.35&71.84&82.10&87.27\\

CSViG \cite{CSVIG}&38.0&203
&84.37&91.52&92.35&82.76&90.57&94.69&87.60&93.39&\textbf{94.48}&69.21&81.80&84.43&80.99&89.32&91.49\\

CLAFA~\cite{CLAFA}&\textbf{14.7}&\textbf{22.0}
&85.21&92.01&92.96&89.74&94.59&95.59&88.19&93.73&91.87&70.12&82.43&83.66&83.32&90.69&91.02\\

\textbf{CLAFA-LWGANet-L2}&16.1&22.1
&\textbf{85.90}&\textbf{92.42}&\textbf{93.25}&\textbf{90.92}&\textbf{95.24}&\textbf{96.51}&\textbf{88.27}&\textbf{93.77}&91.76
&\textbf{70.79}&\textbf{82.90}&\textbf{87.69}&\textbf{83.97}&\textbf{91.08}&\textbf{92.30}\\
\bottomrule
\end{tabular}	}\caption{Change detection experimental results on LEVIR-CD test set~\cite{chen2020spatial}, WHU-CD test set~\cite{ji2018fully}, CDD-CD test set~\cite{lebedev2018change}, and SYSU-CD test set~\cite{shi2021deeply}.} \label{table_CD}
\vspace{-3mm}
\end{table*}

\subsection{Evaluation on Object Detection} \label{detection}

Oriented object detection presents a stringent test for any model's multi-scale capabilities, due to the extreme scale variations where tiny vehicles often co-exist with large airplanes and harbors. We evaluated LWGANet on the challenging DOTA-v1.0/1.5 and DIOR-R datasets to assess the efficacy of our multi-scale design in this demanding context.

We compare against SOTA backbones including ResNet-50~\cite{he2016deep}, ARC-R50~\cite{ARC}, LSKNet-S~\cite{li2023large}, PKINet-S~\cite{cai2024pkinet}, and light-weight backbone networks such as FasterNet-T2~\cite{chen2023run}, EfficientFormerV2-S2~\cite{Eff_Formerv2}, and DecoupleNet-D2~\cite{lu2024decouplenet}.
All backbones are integrated within the Oriented R-CNN~\cite{ORCNN} detector. Parameters ($\#$P), FLOPs, and speeds (FPS) were tested by 1,024$\times$1,024 input size on the RTX 3090 GPU.

As shown in Table \ref{tab_Det_DOTA10_15_DIORR}, LWGANet-L2 establishes a new SOTA among light-weight backbones. It achieved impressive mAP scores of 79.02\% on DOTA-v1.0, 72.91\% on DOTA-v1.5, and 68.53\% on DIOR-R. This performance surpassed even task-specialized, heavier backbones such as PKINet-S. Furthermore, Table \ref{tab_Det_DOTA10_compar} highlights its efficiency: LWGANet was more accurate, used significantly fewer FLOPs (38.8G vs. 70.2G), and was much faster in practice (19.4 FPS vs. 5.4 FPS) than PKINet-S.
This strong performance on datasets with vast scale differences confirms that leveraging channel redundancy through feature decoupling is a highly effective strategy for the multi-scale challenge, outperforming the uniform processing of general-purpose architectures. Concurrently, the model's ability to localize objects accurately within large scenes is supported by the TGFI module's design, which mitigates spatial redundancy for efficient context aggregation. The results suggest that the flexible receptive fields from our LWGA module are beneficial for localizing irregularly shaped objects. Detailed results for each category are presented in the Appendix.

\subsection{Evaluation on Semantic Segmentation} \label{segmentation}
Semantic segmentation demands pixel-level precision, requiring a model to simultaneously recognize fine boundaries and understand broader semantic context. This task directly evaluates the fusion of local and global information within our proposed architecture.

LWGANet was built within the UnetFormer~\cite{unetformer} decoder. The results on UAVid and LoveDA (Table \ref{tab_seg_UAVid_loveda}) confirm the effectiveness of LWGANet. On UAVid, It achieved 69.1\% mIoU, and on the more complex LoveDA dataset, it set a new SOTA with 53.6\% mIoU, all while maintaining a compact size. This strong performance, especially in scenes featuring both small moving objects and large static structures, underscores the capabilities of our multi-pathway design. As visualized in Figure \ref{visual_UAVid}, our model successfully integrates point-level detail with global context, a direct result of the comprehensive representations fused from the four specialized pathways of the LWGA module.

\begin{table*}[t] \centering \footnotesize	
	\renewcommand{\arraystretch}{0.96}	
	\setlength{\tabcolsep}{6pt}
\begin{tabular}{cccccccccccc}
\toprule
\multicolumn{2}{c}{Components} & \multicolumn{4}{c}{Efficiency} & \multicolumn{6}{c}{Performance Metrics (\%) $\uparrow$} \\
\cmidrule(r){1-2} \cmidrule(lr){3-6} \cmidrule(l){7-12}
TGFI & LWGA & Channels & P(M) $\downarrow$ & FLOPs(G) $\downarrow$ & FPS $\uparrow$ & NWPU & DOTA-v1.0 (val) & LoveDA & LEVIR & WHU & CDD \\
\midrule
&  & 24 & 1.92 & 0.234 & 22941 & 94.33 & 67.61 & 48.61 & 82.26 & 85.84 & 86.53 \\
& \ding{51}  & 32 & 1.72 & 0.210 & 6052  & 95.17 & 69.47 & 48.80 & 83.10 & 86.31 & 87.01 \\
\ding{51} & \ding{51} & 32 & 1.72 & 0.188 & 13234 & 95.49 & 70.08 & 49.20 & 82.93 & 86.62 & 86.90 \\
\bottomrule
\end{tabular}
\caption{Ablation study of LWGANet-L0 on its core components. Performance is reported as Top-1 Acc. (\%) for NWPU, mAP (\%) for DOTA-v1.0 (val), mIoU (\%) for LoveDA, and IoU for change detection datasets. } \label{tab:ablation_components_plain}
\vspace{-3mm}
\end{table*}

\subsection{Evaluation on Change Detection} \label{change_detection}

Change detection requires comparing bi-temporal images to identify semantic changes, a task that is highly sensitive to both subtle feature differences and alignment errors. An effective backbone must provide robust, discriminative features to minimize false positives caused by seasonal or lighting variations. As presented in Table \ref{table_CD}, integrating LWGANet as the backbone consistently boosts the performance of SOTA change detection decoders. For instance, A2Net-LWGANet-L0 improved upon the baseline A2Net across all metrics on LEVIR-CD and WHU-CD, while being more parameter-efficient. Similarly, CLAFA-LWGANet-L2 achieved a new SOTA on all four datasets, with a notable 1.18\% IoU gain on WHU-CD. These results highlight that the rich, multi-scale features produced by LWGANet provide a more robust foundation for bi-temporal feature comparison. The model's ability to capture both fine-grained changes and large-scale transformations demonstrates its adaptability and effectiveness for this demanding task.

Across all these RS visual tasks, LWGANet demonstrates exceptional versatility and superior performance, providing an optimal balance between accuracy, parameter efficiency, and computational cost. Notably, LWGANet even surpasses task-specific backbones like LSKNet and PKINet for detection, and RSSFormer for segmentation, in accuracy while maintaining a more light-weight design.

\subsection{Ablation Studies} \label{ablation}
\subsubsection{Quantitative Analyses.}
To validate the individual contributions and synergistic effects of our core components, we conducted a detailed ablation study on the TGFI and LWGA modules using LWGANet-L0. The results, summarized in Table \ref{tab:ablation_components_plain}, confirm that the full model integrating both TGFI and LWGA achieves the highest accuracy on the vast majority of tasks, validating their synergistic benefit. Crucially, the data reveals that TGFI's sparse sampling not only reduces spatial redundancy but also significantly accelerates the heterogeneous LWGA module, leading to an optimal balance of superior accuracy and high practical inference speed.

\subsubsection{Qualitative Analyses.}
To validate our hypothesis that a synergistic, multi-pathway design is superior to any single-paradigm. We constructed several variants of our network, each exclusively using one type of attention module (GPA, RLA, SMA, or SGA) throughout all blocks. These experiments directly compare the feature extraction capabilities of the standalone modules against the full LWGANet, ensuring a fair comparison by utilizing comparable parameters and FLOPs (stem dims set to 64, 64, 96, 96, and 96).

\begin{figure}[t]	\centering	
	\includegraphics[width=1\linewidth]{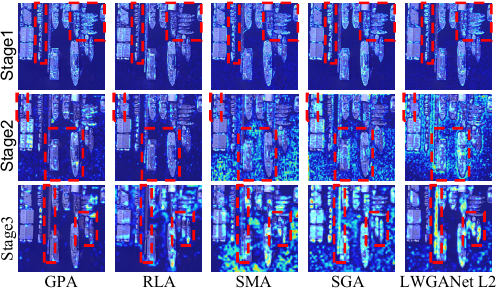}	\vspace{-2mm}
	\caption{The class activation maps (CAMs) visual results. The figure selected from the DOTA 1.0 test set.} \label{cam_visual}
\vspace{-3mm}
\end{figure}

The qualitative results in Figure \ref{cam_visual} provide a visual confirmation of our design's efficacy. The Class Activation Maps (CAMs) for the standalone modules clearly show their inherent biases: GPA focuses on scattered high-frequency points, RLA activates on local textures, while SMA and SGA produce smoother, larger activation regions. In contrast, the CAM for the full LWGANet is both focused and comprehensive. It precisely highlights the target objects while simultaneously capturing their broader context. This visual evidence demonstrates that LWGA module is not just mixing features, but is successfully fusing complementary representations into a coherent and superior whole.

\subsection{Limitations and Future Works}   \label{lim_future}
Future development of LWGANet can be guided by two key objectives: enhancing its architectural adaptivity and improving its practical deployment efficiency.\\
\textbf{(1) Architectural Adaptivity.} LWGANet currently employs a static design with fixed channel partitions and hard-coded hyperparameters. A promising direction is to introduce dynamism, for instance, by using neural architecture search (NAS) to learn optimal channel allocations for each pathway or other structural parameters. This would boost the model's adaptability to diverse data distributions.\\
\textbf{(2) Practical Efficiency.} While the full LWGANet is highly efficient, the heterogeneous operations within the LWGA module can introduce overhead compared to uniform, convolution-only architectures. The TGFI module effectively mitigates this by reducing the computational load on the attention pathways, thereby boosting throughput. However, to unlock maximum performance on edge devices, future work could focus on optimizing the interplay of these diverse operations. This could involve engineering-driven solutions like operator fusion and custom CUDA kernels, or algorithmic approaches to minimize execution divergence and memory access costs.

\vspace{-2mm}
\section{Conclusion}	\label{conclusion}

In this paper, we introduced LWGANet, a light-weight backbone that redefines efficient feature extraction for RS images by systematically addressing the core issues of \textbf{spatial and channel redundancy}. Comprehensive experiments on 12 datasets across four major visual tasks validate the superiority in both accuracy and efficiency. By synergistically resolving these core redundancies, LWGANet establishes a new, robust, and versatile baseline for a wide range of RS applications, particularly on resource-constrained devices.

\vspace{-2mm}
\section{Acknowledgments}
The work was partly supported by the NSFC Key Project of Joint Fund for Enterprise Innovation and Development (U24A20342), Natural Science Foundation of Shanghai (25ZR1402268), and National Natural Science Foundation of China (62576006, 62506229 and 61976004).

\setcounter{page}{1}
\appendix
\onecolumn
\section*{Appendix: Supplementary Materials for LWGANet} \label{Appendix}

This document provides supplementary materials to the main paper, ``LWGANet: Addressing Spatial and Channel Redundancy in Remote Sensing Visual Tasks with Light-Weight Grouped Attention.'' The purpose of this appendix is to offer comprehensive details that ensure the full reproducibility of our work and provide deeper insights into our methodology and findings. We elaborate on architectural specifics, dataset characteristics, implementation protocols, and present additional experimental results that complement the analyses in the main paper.

The appendix is organized as follows:
\begin{itemize}
	\item \textbf{Appendix A} presents a thorough breakdown of the LWGANet architectural configurations.
	\item \textbf{Appendix B} details the 12 public benchmark datasets used in our evaluation.
	\item \textbf{Appendix C} outlines the precise experimental protocols for all downstream tasks.
	\item \textbf{Appendix D} contains extended quantitative results and qualitative visualizations.
	\item \textbf{Appendix E} provides a comprehensive analysis of our ablation studies.
	\item \textbf{Appendix F} discuss a further rationale design.
\end{itemize}

\section{LWGANet Architecture Details} \label{Appendix A}

\begin{figure*}[h]
	\centering
	\includegraphics[width=1\linewidth]{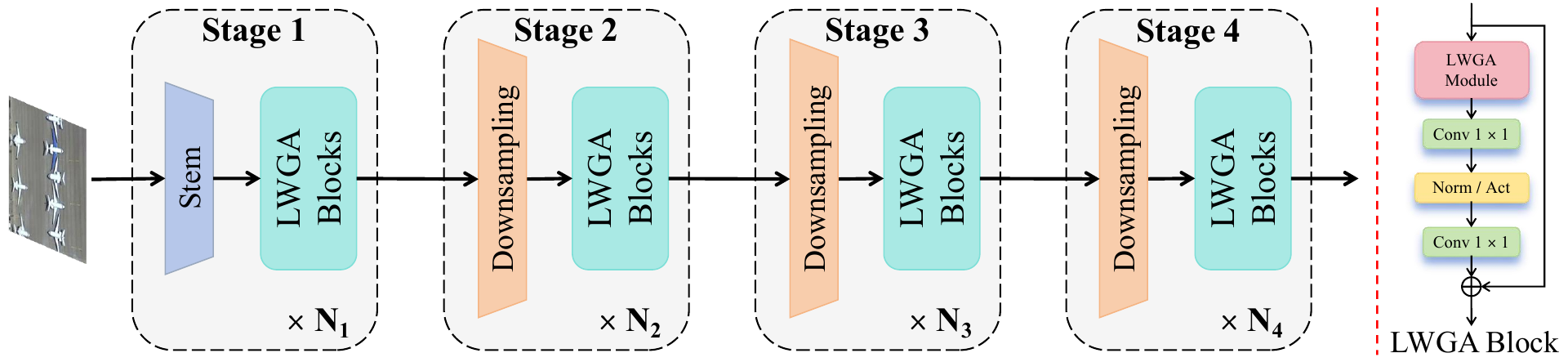}
	\caption{An overview of the LWGANet architecture. The network is organized into four hierarchical stages, producing feature maps with spatial dimensions of $\frac{H}{4} \times \frac{W}{4}$, $\frac{H}{8} \times \frac{W}{8}$, $\frac{H}{16} \times \frac{W}{16}$, and $\frac{H}{32} \times \frac{W}{32}$. The corresponding channel dimensions are $C$, $2C$, $4C$, and $8C$, where $H$, $W$, and $C$ denote the input height, width, and base channel number, respectively.}
	\label{fig_LWGANet}
\end{figure*}

The LWGANet architecture is designed as a lightweight, hierarchical backbone for efficient multi-scale feature extraction in remote sensing (RS) visual tasks. As illustrated in Figure \ref{fig_LWGANet}, the architecture adopts a four-stage pyramidal structure. Each stage progressively reduces the spatial resolution of the feature maps while increasing the channel depth, enabling the network to learn a rich hierarchy of features from local details to global context. To accommodate varying computational budgets and performance requirements, we instantiate LWGANet in three distinct variants: LWGANet-L0, LWGANet-L1, and LWGANet-L2. The detailed configurations for these variants are summarized in Table \ref{tab_achitecture}.

\begin{table*}[h] \centering \small
	\setlength{\tabcolsep}{10pt}
	\renewcommand{\arraystretch}{1.2}
	\begin{tabular}{c|c|c|ccc} \toprule
		\multirow{2}*{Stage} & \multirow{2}*{\makecell[c]{Feature Map \\ Size}} & \multirow{2}*{Layer} & \multicolumn{3}{c}{Channels (In/Out)}\\ \cline{4-6}
		& & & L0 & L1 & L2 \\ \toprule
		\multirow{2}*{1} & \multirow{2}*{$\frac{H}{4}\times \frac{W}{4}$} & \texttt{Stem Layer} & $3/32$ & $3/64$ & $3/96$ \\ \cline{3-6}
		& & \texttt{$\left[LWGA Block\right] \times N_1$} &$32/32$ & $64/64$ & $96/96$ \\ \hline
		\multirow{2}*{2} & \multirow{2}*{$\frac{H}{8}\times \frac{W}{8}$} & \texttt{DRFD Module} & $32/64$ & $64/128$ & $96/192$ \\ \cline{3-6}
		& & \texttt{$\left[LWGA Block\right] \times N_2$} & $64/64$ & $128/128$ & $192/192$ \\ \hline
		\multirow{2}*{3} & \multirow{2}*{$\frac{H}{16}\times \frac{W}{16}$} & \texttt{DRFD Module} & $64/128$ & $128/256$ & $192/384$ \\ \cline{3-6}
		& & \texttt{$\left[LWGA Block\right] \times N_3$} & $128/128$ & $256/256$ & $384/384$ \\  \hline
		\multirow{2}*{4} & \multirow{2}*{$\frac{H}{32}\times \frac{W}{32}$} & \texttt{DRFD Module} & $128/256$ & $256/512$ & $384/768$\\ \cline{3-6}
		& &  \texttt{$\left[LWGA Block\right] \times N_4$} & $256/256$ & $512/512$ & $768/768$ \\ \midrule
		\multicolumn{3}{c|}{SMA distance} & $[11,11,11,11]$ & $[11,11,11,11]$ & $[11,11,11,11]$\\ \hline
		\multicolumn{3}{c|}{Number of Blocks} & $[1, 2, 4, 2]$ & $[1, 2, 4, 2]$ & $[1, 4, 4, 2]$ \\ \hline
		\multicolumn{3}{c|}{Activation} & GELU & GELU & ReLU \\ \hline
		\multicolumn{3}{c|}{Dropout} & $0.0$ & $0.1$ & $0.1$ \\ \hline
		\multicolumn{3}{c|}{Parameters (224$\times$224 input)} & 1.72M & 5.90M & 13.0M \\ 
		\multicolumn{3}{c|}{FLOPs (224$\times$224 input)} & 0.186G & 0.709G & 1.87G \\  \bottomrule
	\end{tabular}
	\caption{Detailed architectural configurations of the LWGANet variants (L0, L1, L2).} \label{tab_achitecture}
\end{table*}

\paragraph{Architectural Blueprint.}
LWGANet's design follows a consistent pattern across its four stages, which operate at spatial resolutions of $\frac{H}{4} \times \frac{W}{4}$, $\frac{H}{8} \times \frac{W}{8}$, $\frac{H}{16} \times \frac{W}{16}$, and $\frac{H}{32} \times \frac{W}{32}$ respectively, where $H$ and $W$ are the height and width.
\begin{itemize}
	\item \textbf{Stage 1} begins with a \texttt{Stem Layer}, which performs an initial 4$\times$ downsampling of the input image and projects it into a higher-dimensional feature space. This is followed by a series of $N_1$ \texttt{LWGA Block}s for initial feature refinement.
	\item \textbf{Stages 2, 3, and 4} each commence with a \texttt{DRFD Module} \cite{lu2023robust}. The DRFD module is responsible for spatial downsampling (by a factor of 2) and channel expansion (typically doubling the channels), mitigating information loss during resolution reduction. Following the DRFD module, a stack of $N_k$ ($k \in \{2, 3, 4\}$) \texttt{LWGA Block}s is employed to enhance the feature representation at the new scale. The \texttt{LWGA Block}, the network's core computational unit, employs a lightweight grouped attention mechanism to capture multi-scale features.
\end{itemize}

\paragraph{Model Variants and Complexity.}
To provide a flexible trade-off between model capacity and computational cost, we define the L0, L1, and L2 variants by adjusting the network's width, depth, and other hyperparameters.
\begin{itemize}
	\item \textbf{Channel Dimensions:} The base channel dimension $C$ is set to 32, 64, and 96 for the L0, L1, and L2 variants, respectively. The channels at each stage scale accordingly, as detailed in Table \ref{tab_achitecture}.
	\item \textbf{Block Depth:} The number of \texttt{LWGA Block}s per stage, denoted by $[N_1, N_2, N_3, N_4]$, is configured as $[1, 2, 4, 2]$ for the L0 and L1 models, and increased to $[1, 4, 4, 2]$ for the larger L2 model to enhance its representational power.
	\item \textbf{Hyperparameters:} The L0 and L1 variants utilize the GELU activation function, while the L2 variant employs ReLU. To manage overfitting in the larger models, a dropout rate of 0.1 is applied to L1 and L2, whereas L0 is trained without dropout.
	\item \textbf{Computational Profile:} These design choices result in a clear progression of model scale. The L0 model is the most compact with 1.72M parameters and 0.186G FLOPs. The L1 model offers a balanced profile with 5.90M parameters and 0.709G FLOPs. The L2 model provides the highest capacity, with 13.0M parameters and 1.87G FLOPs.
\end{itemize}

In summary, the hierarchical design of LWGANet, combined with its specialized DRFD and LWGA modules, facilitates effective and efficient multi-scale feature learning. The availability of three scalable variants makes LWGANet a versatile and powerful solution, adaptable to a wide spectrum of RS applications with diverse resource constraints.

\section{Experimental Datasets}
This section provides a detailed overview of the 12 benchmark datasets employed in our experimental evaluation, spanning four major RS visual tasks: scene classification, oriented object detection, semantic segmentation, and change detection.

\begin{table}[h]\centering \label{tab:dataset_summary}
	\renewcommand{\arraystretch}{1.3}
	\resizebox{\textwidth}{!}{%
		\begin{tabular}{@{}lllrrllc@{}} \toprule
			\textbf{Task} & \textbf{Dataset Name} & \textbf{\# Images/Pairs} & \textbf{\# Classes} & \textbf{\# Instances} & \textbf{Image Size (px)} & \textbf{Spatial Res. (m)} & \textbf{Reference} \\ \midrule
			
			\multirow{3}{*}{\begin{tabular}[c]{@{}l@{}}Scene \\ Classification\end{tabular}} 
			& UCMerced (UCM) & 2,100 & 21 & N/A & 256$\times$256 & 0.3 & \cite{yang2010bag} \\
			& Aerial Image (AID) & 10,000 & 30 & N/A & 600$\times$600 & 0.5 -- 8 & \cite{xia2017aid} \\
			& NWPU-RESISC45 & 31,500 & 45 & N/A & 256$\times$256 & 0.2 -- 30 & \cite{NWPU} \\ \midrule
			
			\multirow{3}{*}{\begin{tabular}[c]{@{}l@{}}Oriented \\ Object \\ Detection \end{tabular}} 
			& DOTA 1.0 & 2,806 & 15 & 188,282 & 800² -- 20,000² & Varies & \cite{dota} \\
			& DOTA 1.5 & 2,806 & 16 & 402,089 & 800² -- 20,000² & Varies & \cite{dota} \\
			& DIOR-R & 23,463 & 20 & 192,472 & 800$\times$800 & 0.5 -- 30 & \cite{AOPG} \\ \midrule
			
			\multirow{2}{*}{\begin{tabular}[c]{@{}l@{}}Semantic \\ Segmentation\end{tabular}} 
			& UAVid & 300 & 8 & N/A & up to 4096$\times$2160 & Varies & \cite{lyu2020uavid} \\
			& LoveDA & 5,987 & 7 & N/A & 1024$\times$1024 & 0.3 & \cite{loveda} \\ \midrule
			
			\multirow{4}{*}{\begin{tabular}[c]{@{}l@{}}Change \\ Detection\end{tabular}} 
			& LEVIR-CD & 637 pairs (10,240 patches) & 2 & N/A & 1024$\times$1024 (cropped) & 0.5 & \cite{chen2020spatial} \\
			& WHU-CD & 1 pair (7,432 patches) & 2 & N/A & Large (cropped) & 0.075 -- 0.5 & \cite{ji2018fully} \\
			& CDD-CD & 11 pairs (16,000 patches) & 2 & N/A & Varies (cropped) & 0.03 -- 1 & \cite{lebedev2018change} \\
			& SYSU-CD & 20,000 pairs & 2 & N/A & 256$\times$256 & 0.5 & \cite{shi2021deeply} \\ \bottomrule
		\end{tabular}%
	}\caption{Summary of the 12 benchmark datasets. The datasets cover four major RS tasks.}
\end{table}

\subsection{Scene Classification Datasets}
\subsubsection{UCMerced Land Use (UCM).} The UCM dataset~\cite{yang2010bag} is a seminal benchmark for RS scene classification. It consists of 2,100 aerial images, each with a size of 256$\times$256 pixels and a spatial resolution of 0.3 meters. The dataset is evenly distributed across 21 land-use categories (e.g., agricultural, residential, forest), with 100 images per category. Sourced from the United States Geological Survey (USGS) National Map, UCM is widely used for initial benchmarking due to its diversity and manageable size.

\subsubsection{Aerial Image Dataset (AID).} AID~\cite{xia2017aid} is a larger-scale dataset for aerial scene classification, comprising 10,000 images across 30 distinct scene types (e.g., airport, forest, viaduct). Each image measures 600$\times$600 pixels, with spatial resolutions varying from 0.5 to 8 meters. Sourced from Google Earth, AID's scale and resolution variability make it a robust benchmark for evaluating the generalization capabilities of deep learning models.

\subsubsection{NWPU-RESISC45.} The NWPU-RESISC45 dataset~\cite{NWPU} is a comprehensive and challenging benchmark for scene classification. It contains 31,500 images distributed among 45 scene classes, with 700 images per class. The images have a resolution of 256$\times$256 pixels and spatial resolutions ranging from 0.2 to 30 meters. Its large scale, high number of classes, and significant intra-class diversity make it an ideal testbed for modern deep learning architectures.

\subsection{Oriented Object Detection Datasets}
\subsubsection{DOTA 1.0.} The DOTA 1.0 dataset~\cite{dota} is a large-scale benchmark for object detection in aerial images. It contains 2,806 high-resolution images (ranging from 800$\times$800 to 20,000$\times$20,000 pixels) and 188,282 object instances across 15 categories. Objects are annotated with oriented bounding boxes (OBB) using arbitrary quadrilaterals, making it a standard for evaluating oriented object detection algorithms. The 15 object categories include: Plane (PL), Baseball diamond (BD), Bridge (BR), Ground track field (GTF), Small vehicle (SV), Large vehicle (LV), Ship (SH), Tennis court (TC), Basketball court (BC), Storage tank (ST), Soccer-ball field (SBF), Roundabout (RA), Harbor (HA), Swimming pool (SP), and Helicopter (HC).

\subsubsection{DOTA 1.5.} DOTA 1.5~\cite{dota} is an updated version of DOTA 1.0, using the same images but with revised annotations. It addresses challenges such as small object detection by adding numerous new instances, bringing the total to 402,089. It also introduces a new category, "container crane (CC)," increasing the total to 16.

\subsubsection{DIOR-R.} The DIOR-R dataset~\cite{AOPG} is an extension of the DIOR dataset, specifically tailored for oriented object detection. It comprises 23,463 images and 192,472 instances across 20 object categories, with spatial resolutions from 0.5 to 30 meters. All objects are annotated with rotated bounding boxes, providing a rich resource for developing and testing robust oriented detectors. The 20 common object categories include: Airplane (APL), Airport (APO), Baseball field (BF), Basketball court (BC), Bridge (BR), Chimney (CH), Expressway service area (ESA), Expressway toll station (ETS), Dam (DAM), Golf field (GF), Ground track field (GTF), Harbor (HA), Overpass (OP), Ship (SH), Stadium (STA), Storage tank (STO), Tennis court (TC), Train station (TS), Vehicle (VE) and Windmill (WM).

\subsection{Semantic Segmentation Datasets}
\subsubsection{UAVid.} The UAVid dataset~\cite{lyu2020uavid} is designed for high-resolution semantic segmentation from Unmanned Aerial Vehicle (UAV) platforms. It includes 30 video sequences, from which 300 images are densely annotated with pixel-level labels. The images feature resolutions up to 4,096$\times$2,160 pixels and cover eight semantic classes (e.g., buildings, roads, moving cars), presenting challenges related to large scale variations and oblique viewing angles.

\subsubsection{LoveDA.} The LoveDA dataset~\cite{loveda} targets land-cover mapping and is notable for its focus on domain adaptation between urban and rural scenes. It contains 5,987 images of 1,024$\times$1,024 pixels with a 0.3-meter spatial resolution. The dataset is annotated with seven land-cover classes and is split into urban and rural subsets, making it ideal for evaluating cross-domain segmentation performance.

\subsection{Change Detection Datasets}
\subsubsection{LEVIR-CD.} The LEVIR-CD dataset~\cite{chen2020spatial} is a large-scale benchmark for building change detection. It consists of 637 pairs of high-resolution (1,024$\times$1,024 pixels, 0.5-meter resolution) bi-temporal images from Google Earth. For standardized evaluation, the images are cropped into 256$\times$256 non-overlapping patches, yielding 7,120 training, 1,024 validation, and 2,048 test pairs.

\subsubsection{WHU-CD.} The WHU-CD dataset~\cite{ji2018fully} focuses on building change detection in very high-resolution aerial imagery (0.075--0.5 meters). It contains a single large image pair covering a region that underwent significant urban development, which is cropped into 7,432 non-overlapping patch pairs for analysis.

\subsubsection{CDD-CD.} The CDD-CD dataset~\cite{lebedev2018change} provides 11 pairs of season-varying satellite images with resolutions from 0.03 to 1 meter per pixel. It captures changes in both man-made objects and natural landscapes. The dataset is pre-processed into 256$\times$256 patches, resulting in 10,000 training, 3,000 validation, and 3,000 test pairs.

\subsubsection{SYSU-CD.} The SYSU-CD dataset~\cite{shi2021deeply} contains 20,000 pairs of 256$\times$256 pixel images with a 0.5-meter resolution, focusing on urban changes in Hong Kong. It is divided into 12,000 training, 4,000 validation, and 4,000 test pairs, providing a large-scale resource for training data-hungry change detection models.

\section{Experimental Setup}
This section details the experimental protocols, including training configurations, data processing, and evaluation environments for each downstream task.

\subsection{Classification Experimental Setup}
For scene classification, all models were trained from scratch for 300 epochs to assess their intrinsic learning capabilities without external pre-training. The datasets were partitioned into training and validation sets using an 80/20 split. Key training parameters are listed below:
\begin{itemize}
	\item \textbf{Input Resolution:} All images were resized to 224$\times$224 pixels.
	\item \textbf{Optimizer:} We used the AdamW optimizer~\cite{loshchilovdecoupled} with a learning rate of $1 \times 10^{-4}$ and a weight decay of $5 \times 10^{-2}$.
	\item \textbf{Learning Rate Schedule:} A cosine decay schedule~\cite{cosine} was employed, preceded by a 2-epoch linear warmup phase with an initial warmup factor of $1 \times 10^{-3}$.
	\item \textbf{Batch Size:} A batch size of 64 was used for all experiments.
	\item \textbf{Data Augmentation:} Standard augmentation techniques were applied, including RandomResizedCrop, RandomHorizontalFlip, and RandAugment~\cite{cubuk2020randaugment}.
	\item \textbf{Loss Function:} The standard cross-entropy loss was used for optimization.
	\item \textbf{Environment:} Experiments were conducted on an NVIDIA RTX 3090 GPU using the PyTorch framework with Ubuntu20.04. Inference speeds were benchmarked on three platforms: NVIDIA RTX 3090 (GPU), Intel i9-11900K (CPU), and NVIDIA AGX-XAVIER (ARM).
\end{itemize}

\subsection{Detection Experimental Setup}
For oriented object detection, we adopted the Oriented R-CNN~\cite{ORCNN} framework implemented in MMRotate~\cite{zhou2022mmrotate}. All backbone models were pre-trained on ImageNet-1k~\cite{imagenet} for 300 epochs.
\begin{itemize}
	\item \textbf{Data Preprocessing:} For DOTA 1.0 and DOTA 1.5, original images were cropped into 1,024$\times$1,024 patches with a 200-pixel overlap. For DIOR-R, the input size was 800$\times$800.
	\item \textbf{Training Schedule:} Models were trained for 36 epochs using the AdamW optimizer with an initial learning rate of $2 \times 10^{-4}$ and a weight decay of 0.05. The batch size was set to 8.
	\item \textbf{Data Augmentation:} Random resizing and flipping were applied during training to enhance model robustness.
	\item \textbf{Evaluation:} Following standard evaluation protocols, we trained models on the trainval splits and reported performance on the test splits.
\end{itemize}

\subsection{Segmentation Experimental Setup}
For semantic segmentation, we employed UnetFormer~\cite{unetformer} as the segmentation head. The backbones were pre-trained on ImageNet-1k.
\begin{itemize}
	\item \textbf{UAVid Dataset:} Models were trained for 30 epochs with a batch size of 8. Input images were resized to 1,024$\times$1,024. Augmentations included random vertical/horizontal flips and random brightness adjustments. Test-time augmentation (TTA) with flips was used for evaluation.
	\item \textbf{LoveDA Dataset:} Models were trained for 70 epochs with a batch size of 16. Images were randomly cropped to 1,024$\times$1,024. Augmentations included random scaling, flips, and rotations. Multi-scale and flip augmentations were used during testing.
	\item \textbf{Optimizer:} The AdamW optimizer was used with a base learning rate of $6 \times 10^{-4}$, adjusted by a cosine annealing schedule.
\end{itemize}

\subsection{Change Detection Experimental Setup}
For change detection, we integrated our backbone into two different decoders, A2Net~\cite{A2Net} and CLAFA~\cite{CLAFA}, following their respective training protocols. Backbones were pre-trained on ImageNet-1k.
\begin{itemize}
	\item \textbf{Optimizer:} The Adam optimizer was used with a momentum of 0.9 and a weight decay of $1 \times 10^{-4}$.
	\item \textbf{Learning Rate Schedule:} The learning rate was initialized to $1 \times 10^{-3}$ and decayed to zero over 20,000 iterations using a polynomial schedule.
	\item \textbf{Batch Size:} A batch size of 64 was used.
	\item \textbf{Model Selection:} The model weights that achieved the highest F1 score on the validation set were selected for final evaluation on the test set.
\end{itemize}

\clearpage

\begin{table*}[h] \centering	\small
	\renewcommand{\arraystretch}{1.3}
	\setlength{\tabcolsep}{3mm}{
		\begin{tabular}{cccc|ccc|ccc} \toprule
			\multirow{2}{*}{Method} & \multirow{2}{*}{\makecell[c]{Backbone \\ Type}} & \multirow{2}{*}{\makecell[c]{Params. \\ (M) $\downarrow$}} & \multirow{2}{*}{\makecell[c]{FLOPs \\ (G) $\downarrow$}} & \multicolumn{3}{c}{Top-1 Accuracy (\%) $\uparrow$} & \multicolumn{3}{c}{Speed (FPS) $\uparrow$} \\ \cline{5-10}
			& & & & NWPU & AID & UCM & GPU & CPU & ARM \\ \midrule
			MobileNet V2 1.0$\times$ & CNN & 2.28 & 0.319 & 95.06 & 93.65 & 97.14 & 11301 & 49.11 & 785.4 \\
			FasterNet T0 & CNN & 2.68 & 0.338 & 93.30 & 92.85 & 94.75 & 18276 & 106.4 & 839.7 \\
			EdgeViT XXS & Transformer & 3.79 & 0.546 & 94.75 & 93.10 & 95.24 & 4153 & 46.36 & 259.9 \\
			EfficientformerV2 S0 & Transformer & 3.36 & 0.396 & 94.52 & 93.80 & 97.14 & 1299 & 54.00 & 272.0 \\
			$^{\star}$MobileViT XXS & Hybrid & 1.03 & 0.333 & 94.37 & 93.30 & 95.00 & 4811 & 31.87 & 453.6 \\
			\rowcolor{gray!30}\textbf{LWGANet L0} & Hybrid & 1.72 & 0.186 & \textbf{95.49} & \textbf{94.60} & \textbf{98.57} & \textbf{13234} & \textbf{80.00} & \textbf{687.8} \\
			\midrule
			MobileNet V2 2.0$\times$ & CNN & 8.81 & 1.17 & 95.35 & 93.85 & 97.86 & 5567 & 17.03 & 372.3 \\
			FasterNet T1 & CNN & 6.37 & 0.855 & 93.73 & 93.20 & 94.52 & 11876 & 51.42 & 650.1 \\
			PVT V2 B0 & Transformer & 3.42 & 0.533 & 94.35 & 93.10 & 96.43 & 3843 & 40.26 & 243.4 \\
			EfficientformerV2 S1 & Transformer & 5.87 & 0.661 & 94.97 & 93.95 & 96.90 & 1211 & 36.96 & 204.5 \\
			$^{\star}$MobileViT XS & Hybrid & 2.02 & 0.900 & 94.90 & 95.20 & 96.43 & 3300 & 12.99 & 306.3 \\
			\rowcolor{gray!30}\textbf{LWGANet L1} & Hybrid & 5.90 & 0.709 & \textbf{95.70} & \textbf{94.85} & \textbf{98.81} & \textbf{6418} & \textbf{34.08} & \textbf{375.8} \\
			\midrule
			FasterNet T2 & CNN & 13.8 & 1.91 & 95.11 & 93.60 & 94.29 & 6852 & 26.40 & 669.8 \\
			PVT V2 B1 & Transformer & 13.5 & 2.04 & 94.62 & 93.45 & 95.71 & 2369 & 15.96 & 145.3 \\
			EfficientformerV2 S2 & Transformer & 12.3 & 1.26 & 95.14 & 94.20 & 97.38 & 642 & 24.74 & 123.8 \\
			$^{\star}$MobileViT S & Hybrid & 5.03 & 1.75 & 95.19 & 95.25 & 97.14 & 2681 & 10.20 & 152.7 \\
			\rowcolor{gray!30}\textbf{LWGANet L2} & Hybrid & 13.0 & 1.87 & \textbf{96.17} & \textbf{95.45} & \textbf{98.57} & \textbf{3308} & \textbf{16.18} & \textbf{274.3} \\
			\bottomrule
	\end{tabular}}
	\caption{Comprehensive comparison on the NWPU, AID, and UCM classification datasets. `$\star$' indicates a training image size of 256$\times$256. FPS data are from RTX 3090 (GPU), Intel i9-11900K (CPU), and NVIDIA AGX-XAVIER (ARM) platforms.} \label{table_cla_sup}
\end{table*}

\section{Quantitative Experimental Results}
This section presents a detailed quantitative analysis of LWGANet's performance across all evaluated tasks, benchmarked against state-of-the-art methods.

\subsection{Classification Experimental Results}
\begin{figure*}[h!]
	\centering
	\includegraphics[width=1\linewidth]{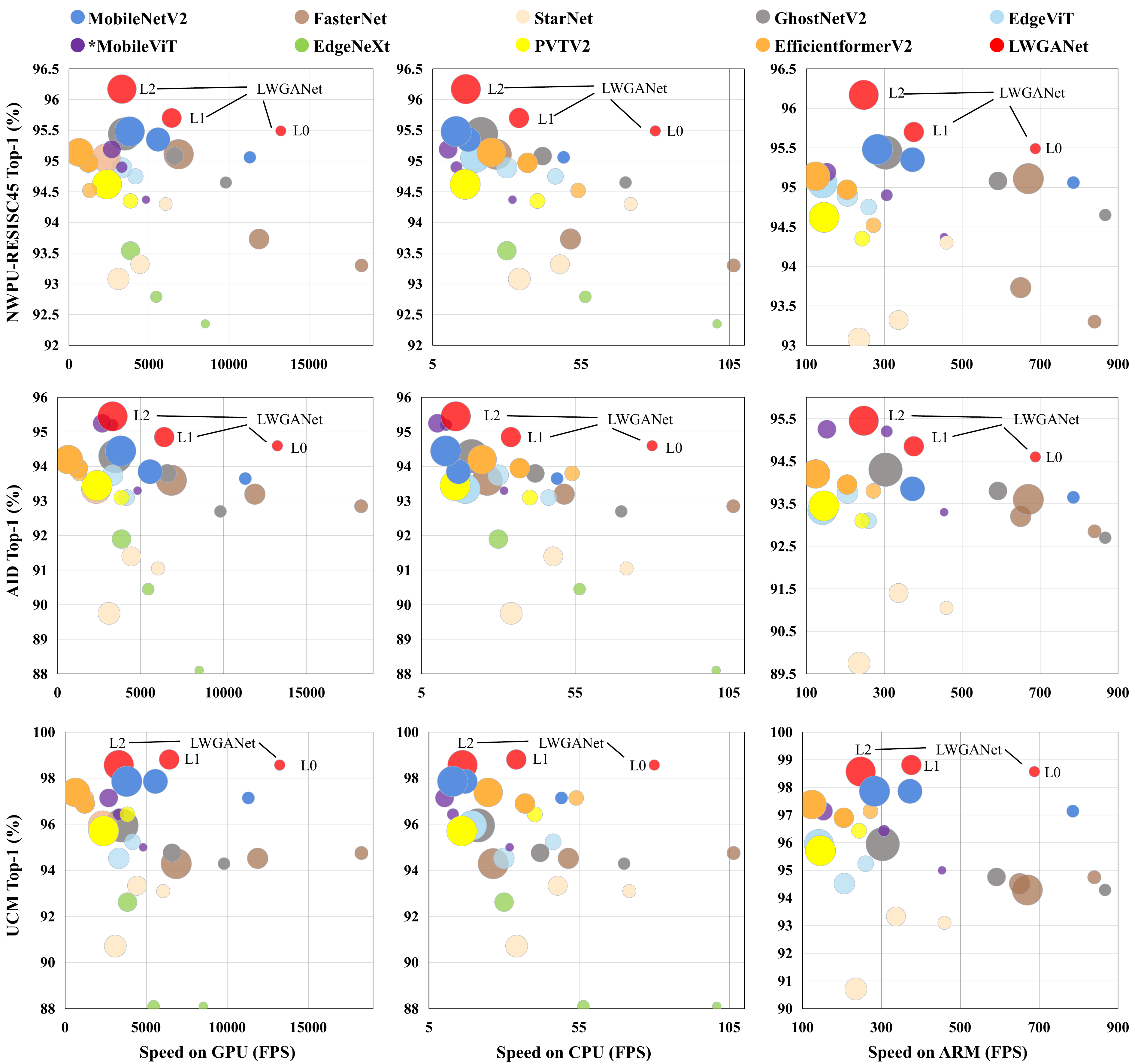}
	\caption{Comparison of Top-1 accuracy, inference speed (FPS), and model parameters on the NWPU, AID, and UCM datasets. The area of each circle is proportional to the model's parameter count. FPS was measured on NVIDIA RTX 3090 (GPU), Intel i9-11900K (CPU), and NVIDIA AGX-XAVIER (ARM) platforms with batch sizes of 256, 16, and 32, respectively. LWGANet (\textcolor{red}{red} circles) consistently achieves the best trade-off.}
	\label{bubble_cla}
\end{figure*}

As demonstrated in Figure \ref{bubble_cla} and detailed in Table \ref{table_cla_sup}, our proposed LWGANet redefines the state-of-the-art trade-off between classification accuracy, inference speed, and model efficiency across the NWPU, AID, and UCM datasets. We conduct a comprehensive comparison against a suite of prominent lightweight models, including CNN-based architectures like MobileNetV2 \cite{mobilenetv2} and FasterNet \cite{chen2023run}, Transformer-based models such as PVT V2 \cite{pvtv2}, EdgeViT \cite{pan2022edgevits}, and EfficientformerV2 \cite{Eff_Formerv2}, as well as hybrid models like MobileViT \cite{mobilevit}. Across all benchmarks, LWGANet consistently delivers a superior performance profile.

\paragraph{Accuracy.} LWGANet consistently outperforms its peers in Top-1 accuracy. For instance, the most compact variant, LWGANet-L0 (1.72M parameters), achieves impressive accuracies of 95.49\%, 94.60\%, and 98.57\% on the NWPU, AID, and UCM datasets, respectively. This performance surpasses not only smaller models like EdgeViT-XXS but also larger competitors such as MobileNetV2-1.0$\times$. This performance advantage is consistent across model scales; the LWGANet-L2 model reaches a remarkable 96.17\% on NWPU, outclassing models with similar or greater parameter counts, including PVTV2-B1, while maintaining a competitive model size.

\paragraph{Inference Speed and Efficiency.} LWGANet demonstrates exceptional throughput across diverse hardware platforms (GPU, CPU, and ARM). The LWGANet-L0 variant achieves 13,234 FPS on an RTX 3090 GPU and 80.00 FPS on an Intel i9-11900K CPU—speeds that are highly competitive and often superior to other models in its class. Even when scaled to the 13.0M-parameter LWGANet-L2, the model maintains excellent efficiency. Its GPU speed of 3,308 FPS is comparable to that of other models in its computational class but is achieved with significantly higher accuracy, highlighting the efficiency of our architecture.

\paragraph{Performance-Efficiency Trade-off.} The bubble plot in Figure \ref{bubble_cla} visually summarizes LWGANet's dominance. The LWGANet series (red circles) consistently occupies the Pareto frontier, achieving a superior balance of accuracy, latency, and model complexity. It avoids the common compromise of sacrificing accuracy for speed or vice-versa, providing a more optimal solution than models that are either faster but less accurate (e.g., FasterNet) or more accurate but computationally heavier (e.g., some Transformer-based models).

\subsection{Object Detection Experimental Results}
\subsubsection{DOTA 1.0 Results.}
Table \ref{Det-dota10} presents a comprehensive comparison of LWGANet-L2 against state-of-the-art oriented object detection methods on the DOTA 1.0 benchmark. We evaluate against a diverse set of approaches, including established detectors like SCRDet \cite{Scrdet}, RoI Transformer \cite{ROI}, Gliding Vertex \cite{xu2020gliding}, and ReDet \cite{redet}. We also benchmark against various backbones integrated within the Oriented R-CNN \cite{ORCNN} framework, such as the classic ResNet-50 \cite{he2016deep} and more recent lightweight architectures like EfficientFormerV2-S2 \cite{Eff_Formerv2}, ARC-R50 \cite{ARC}, LSKNet-S \cite{li2023large}, DecoupleNet-D2 \cite{lu2024decouplenet}, and PKINet-S \cite{cai2024pkinet}. As shown in the table, our model, integrated into the Oriented R-CNN framework, achieves a new state-of-the-art mean Average Precision (mAP) of \textcolor{red}{79.02\%}. This result surpasses all competing methods.

\begin{table*}[h]	\centering	\scriptsize
	\renewcommand{\arraystretch}{1.2}
	\setlength{\tabcolsep}{0.96mm}{
		\begin{tabular}{c|c|cc|c|ccccccccccccccc} \toprule
			\multirow{2}{*}{Methods} & \multirow{2}{*}{Backbone} & \multirow{2}{*}{\makecell[c]{Params. \\ (M) $\downarrow$}} & \multirow{2}{*}{\makecell[c]{FLOPs \\ (G) $\downarrow$}} & \multirow{2}{*}{\makecell[c]{mAP \\ (\%) $\uparrow$}} & \multicolumn{15}{c}{AP per category (\%) $\uparrow$} \\ \cline{6-20}
			& & & & & PL & BD & BR & GTF & SV & LV & SH & TC & BC & ST & SBF & RA & HA & SP & HC \\ \hline
			SCRDet & ResNet-50 & 41.9 & - & 72.61 & 89.98 & 80.65 & 52.09 & 68.36 & 68.36 & 60.32 & 72.41 & 90.85 & \textcolor{red}{87.94} & \textcolor{red}{86.86} & 65.02 & 66.68 & 66.25 & 68.24 & 65.21 \\
			RoI Transformer & ResNet-50 & 55.1 & 225.3 & 74.05 & 89.01 & 77.48 & 51.64 & 72.07 & 74.43 & 77.55 & 87.76 & 90.81 & 79.71 & 85.27 & 58.36 & 64.11 & 76.50 & 71.99 & 54.06 \\
			Gliding Vertex & ResNet-50 & 41.1 & 211.3 & 75.02 & 89.64 & 85.00 & 52.26 & \textcolor{blue}{77.34} & 73.01 & 73.14 & 86.82 & 90.74 & 79.02 & \textcolor{blue}{86.81} & 59.55 & \textcolor{red}{70.91} & 72.94 & 70.86 & 57.32 \\
			ReDet & ReResNet & 31.6 & - & 76.25 & 88.79 & 82.64 & 53.97 & 74.00 & 78.13 & 84.06 & 88.04 & \textcolor{blue}{90.89} & \textcolor{blue}{87.78} & 85.75 & 61.76 & 60.39 & 75.96 & 68.07 & 63.59 \\ \midrule
			\multirow{6}{*}{\makecell[c]{Oriented R-CNN}} & ResNet-50 & 41.1 & 211.4 & 75.87 & 89.46 & 82.12 & 54.78 & 70.86 & 78.93 & 83.00 & 88.20 & \textcolor{red}{90.90} & 87.50 & 84.68 & 63.97 & 67.69 & 74.94 & 68.84 & 52.28 \\
			& $^*$E-FormerV2-S2 & \textcolor{blue}{29.2} & \textcolor{blue}{145.1} & 76.70 & 89.55 & 84.12 & 53.39 & 74.40 & \textcolor{blue}{80.70} & \textcolor{blue}{84.84} & 87.92 & \textcolor{blue}{90.89} & 87.44 & 84.47 & 60.88 & 67.43 & 77.63 & 67.62 & 59.16 \\
			& ARC-R50 & 74.4 & 212.0 & 77.35 & 89.40 & 82.48 & 55.33 & 73.88 & 79.37 & 84.05 & 88.06 & \textcolor{red}{90.90} & 86.44 & 84.83 & 63.63 & \textcolor{blue}{70.32} & 74.29 & 71.91 & 65.43 \\
			& LSKNet-S & 31.0 & 161.0 & 77.49 & \textcolor{blue}{89.66} & \textcolor{red}{85.52} & \textcolor{red}{57.72} & 75.70 & 74.95 & 78.69 & 88.24 & 90.88 & 86.79 & 86.38 & \textcolor{blue}{66.92} & 63.77 & \textcolor{blue}{77.77} & \textcolor{red}{74.47} & 64.82 \\
			& DecoupleNet-D2 & \textcolor{red}{23.3} & \textcolor{red}{142.4} & 78.04 & 89.37 & 83.25 & 54.29 & 75.51 & 79.83 & 84.82 & \textcolor{red}{88.49} & \textcolor{blue}{90.89} & 87.19 & 86.23 & 66.07 & 65.53 & 77.23 & 72.34 & \textcolor{blue}{69.62} \\
			& PKINet-S & 30.8 & 184.6 & \textcolor{blue}{78.39} & \textcolor{red}{89.72} & 84.20 & \textcolor{blue}{55.81} & \textcolor{red}{77.63} & 80.25 & 84.45 & 88.12 & 90.88 & 87.57 & 86.07 & 66.86 & 70.23 & 77.47 & \textcolor{blue}{73.62} & 62.94 \\
			\rowcolor{gray!30} & \textbf{LWGANet-L2} & \textcolor{blue}{29.2} & 159.1 & \textcolor{red}{79.02} & 89.49 & \textcolor{blue}{85.48} & 54.93 & 77.12 & \textcolor{red}{81.59} & \textcolor{red}{85.64} & \textcolor{blue}{88.43} & 90.85 & 87.23 & 86.78 & \textcolor{red}{67.47} & 65.06 & \textcolor{red}{78.23} & 73.33 & \textcolor{red}{73.66} \\ \bottomrule
	\end{tabular}}
	\caption{Comparison with state-of-the-art oriented object detectors on the \textbf{DOTA 1.0 test set}. All methods are evaluated under a single-scale training and testing protocol \textcolor{red}{Red} and \textcolor{blue}{blue} denote the top-two performance in each column. $^*$E-FormerV2-S2 refers to EfficientFormerV2-S2.} \label{Det-dota10}
\end{table*}

Notably, LWGANet-L2 achieves this superior accuracy while maintaining high computational efficiency. With \textcolor{blue}{29.2M} parameters and 159.1 GFLOPs, it is more efficient than PKINet-S and significantly more accurate than models with lower complexity, such as DecoupleNet-D2 and EfficientFormerV2-S2. This demonstrates LWGANet-L2's exceptional ability to balance accuracy and resource consumption. In per-category performance, LWGANet-L2 excels on challenging classes like small vehicle (SV), large vehicle (LV), and helicopter (HC), underscoring the robustness of its feature representation.

\subsubsection{DOTA 1.5 Results.}
On the more challenging DOTA 1.5 test set, as detailed in Table \ref{tab_Det_DOTA15_sup}, we compare LWGANet-L2 against several state-of-the-art backbones within the Oriented R-CNN framework. These include ResNet-50 \cite{he2016deep}, ARC-R50 \cite{ARC}, LSKNet-S \cite{li2023large}, DecoupleNet-D2 \cite{lu2024decouplenet}, PKINet-S \cite{cai2024pkinet}, and EfficientFormerV2-S2 \cite{Eff_Formerv2}. LWGANet-L2 continues to demonstrate its superiority, achieving a new state-of-the-art mAP of \textcolor{red}{72.91\%}. This performance surpasses all other evaluated backbones. The consistent high performance across all 16 categories, without significant drops on any particular class, underscores the robustness of our model. This balanced accuracy profile underscores the effectiveness of our multi-scale attention mechanism, which effectively captures both fine-grained details and global context, a critical capability for handling the diverse object scales and dense scenes present in DOTA 1.5.

\begin{table*}[h] \centering
	\setlength\tabcolsep{5pt}
	\renewcommand\arraystretch{1.3}
	\resizebox{\textwidth}{!}{
		\begin{tabular}{r||cccccccccccccccc||c}
			\rowcolor[rgb]{0.92,0.92,0.92} \textbf{Method} & \textbf{PL} & \textbf{BD} & \textbf{BR} & \textbf{GTF} & \textbf{SV} & \textbf{LV} & \textbf{SH} & \textbf{TC} & \textbf{BC} & \textbf{ST} & \textbf{SBF} & \textbf{RA} & \textbf{HA} & \textbf{SP} & \textbf{HC} & \textbf{CC} & \textbf{mAP $\uparrow$} \\ \hline\hline
			ResNet-50 & 79.95 & 81.00 & 53.90 & 70.59 & \textcolor{blue}{52.48} & 76.21 & 86.98 & \textcolor{blue}{90.88} & 78.33 & 68.26 & 58.94 & 72.60 & 72.75 & 65.32 & 58.18 & 3.72 & 66.88 \\
			ARC-R50 & 80.27 & 82.40 & 54.57 & 73.03 & 52.37 & 80.28 & 87.93 & \textcolor{blue}{90.88} & \textcolor{red}{83.33} & \textcolor{blue}{69.18} & 57.37 & 72.35 & 71.97 & 65.40 & 68.35 & 3.24 & 68.31 \\
			LSKNet-S & 72.05 & \textcolor{blue}{84.94} & \textcolor{blue}{55.41} & \textcolor{red}{74.93} & 52.42 & 77.45 & 81.17 & 90.85 & 79.44 & 69.00 & 62.10 & \textcolor{red}{73.72} & \textcolor{red}{77.49} & \textcolor{red}{75.29} & 55.81 & 42.19 & 70.26 \\
			DecoupleNet-D2 & \textcolor{red}{80.35} & 82.36 & 54.00 & 73.00 & 52.30 & \textcolor{blue}{81.41} & 88.31 & \textcolor{red}{90.89} & 80.22 & \textcolor{red}{69.27} & 58.50 & 68.41 & 75.99 & 70.93 & 72.92 & 39.54 & 71.15 \\
			PKINet-S & \textcolor{blue}{80.31} & \textcolor{red}{85.00} & \textcolor{red}{55.61} & \textcolor{blue}{74.38} & 52.41 & 76.85 & 88.38 & 90.87 & 79.04 & 68.78 & \textcolor{red}{67.47} & 72.45 & 76.24 & 74.53 & 64.07 & 37.13 & 71.47 \\
			$^*$E-FormerV2-S2 & 80.22 & 82.44 & 51.78 & 72.85 & 52.26 & 77.82 & \textcolor{blue}{88.43} & 90.87 & 79.57 & 68.19 & 62.46 & 70.15 & \textcolor{blue}{77.21} & 71.92 & \textcolor{red}{76.37} & \textcolor{red}{52.43} & \textcolor{blue}{72.19} \\
			\hline\hline
			\textbf{LWGANet-L2 (ours)} & 80.00 & 84.40 & 55.31 & 74.10 & \textcolor{red}{52.46} & \textcolor{red}{82.26} & \textcolor{red}{88.84} & 90.86 & 79.34 & 69.03 & \textcolor{blue}{65.47} & 72.06 & 76.98 & \textcolor{blue}{74.74} & \textcolor{blue}{75.24} & \textcolor{blue}{45.40} & \textcolor{red}{72.91} \\ \hline
	\end{tabular}}
	\caption{Experimental results on the DOTA 1.5 test set. All backbones are evaluated within the Oriented R-CNN detector.} \label{tab_Det_DOTA15_sup}
\end{table*}

\subsection{Semantic Segmentation Experimental Results}

\subsubsection{UAVid Results.}
On the UAVid dataset, as detailed in Table \ref{seg-UAVid}, we compare LWGANet-L2 against a variety of state-of-the-art lightweight semantic segmentation methods. These include CoaT \cite{CoaT}, SegFormer \cite{segformer}, and several backbones integrated with the UnetFormer head \cite{unetformer}, such as ResNet18 \cite{he2016deep}, EfficientFormerV2-S2 \cite{Eff_Formerv2}, FasterNet-T2 \cite{chen2023run}, and DecoupleNet-D2 \cite{lu2024decouplenet}. Our LWGANet-L2 backbone achieves a state-of-the-art mean Intersection-over-Union (mIoU) of \textcolor{red}{69.1\%}. This result surpasses all lightweight competitors. While maintaining a competitive model size (12.6M parameters) and inference speed (67.3 FPS), LWGANet-L2 delivers a significant accuracy improvement of +1.3\% mIoU over the strong ResNet18 baseline. It also outperforms other modern lightweight backbones like DecoupleNet-D2 and FasterNet-T2 by a substantial margin. The superior performance is consistent across most categories, with LWGANet-L2 achieving the highest IoU for five of the eight classes, including challenging ones like \textit{Moving Car} and \textit{Human}, demonstrating its powerful feature extraction capabilities.

\begin{table*}[h]	\centering	\scriptsize
	\renewcommand{\arraystretch}{1.2}
	\setlength{\tabcolsep}{1.5mm}{
		\begin{tabular}{c|c|cccc|c|ccccccc} \toprule
			\multirow{2}{*}{Methods} & \multirow{2}{*}{Backbone} & \multirow{2}{*}{\makecell[c]{Params. \\ (M) $\downarrow$}} & \multirow{2}{*}{\makecell[c]{FLOPs \\ (G) $\downarrow$}} & \multirow{2}{*}{\makecell[c]{Speed \\ (FPS) $\uparrow$}} & \multirow{2}{*}{\makecell[c]{mIoU \\ (\%) $\uparrow$}} & \multicolumn{8}{c}{IoU per category (\%) $\uparrow$} \\ \cline{7-14}
			& & & & & & Clutter & Building & Road & Tree & Vegetation & Moving Car & Static Car & Human \\ \midrule
			CoaT & CoaT-Mini & \textcolor{blue}{11.1} & 104.8 & 10.6 & 65.8 & \textcolor{red}{69.0} & \textcolor{red}{88.5} & 80.0 & 79.3 & 62.0 & 70.0 & \textcolor{red}{59.1} & 18.9 \\
			SegFormer & MiT-B1 & 13.7 & 63.3 & 31.3 & 66.0 & 66.6 & 86.3 & 80.1 & 79.6 & 62.3 & 72.5 & 52.5 & 28.5 \\ \midrule
			\multirow{5}{*}{UnetFormer} & EfficientFormerV2-S2 & 12.7 & \textcolor{blue}{35.8} & 54.1 & 65.2 & 63.8 & 82.5 & 78.9 & 78.0 & 63.0 & 73.5 & 51.7 & 29.9 \\
			& FasterNet-T2 & 13.3 & 49.1 & \textcolor{red}{198.8} & 65.7 & 65.3 & 86.2 & 79.6 & 78.8 & 62.1 & 70.9 & 54.8 & 28.1 \\
			& DecoupleNet-D2 & \textcolor{red}{6.8} & \textcolor{red}{32.1} & - & 65.8 & 65.1 & 85.4 & 80.6 & 78.8 & 62.1 & \textcolor{blue}{74.1} & 49.7 & 30.8 \\
			& ResNet18 & 11.7 & 46.9 & 115.6 & \textcolor{blue}{67.8} & \textcolor{blue}{68.4} & 87.4 & \textcolor{blue}{81.5} & \textcolor{blue}{80.2} & \textcolor{blue}{63.5} & 73.6 & 56.4 & \textcolor{blue}{31.0} \\
			\rowcolor{gray!30} & \textbf{LWGANet-L2} & 12.6 & 50.3 & 67.3 & \textcolor{red}{69.1} & \textcolor{red}{69.0} & \textcolor{blue}{87.9} & \textcolor{red}{81.9} & \textcolor{red}{80.5} & 64.6 & \textcolor{red}{76.7} & \textcolor{blue}{59.7} & \textcolor{red}{32.7} \\ \bottomrule
	\end{tabular}}
	\caption{Segmentation results on the UAVid test set. Speeds were evaluated on 1,024$\times$1,024 inputs using an RTX 3090 GPU.} \label{seg-UAVid}
\end{table*}

\subsubsection{LoveDA Results.}
On the LoveDA benchmark, as shown in Table \ref{seg-LoveDA}, we compare UnetFormer with an LWGANet-L2 backbone against several state-of-the-art models, including FactSeg \cite{factseg}, FarSeg \cite{FarSeg}, UnetFormer \cite{unetformer} (with a ResNet50 \cite{he2016deep} backbone), RSSFormer \cite{rssformer}, and LoveNAS \cite{lovenas}. Our configuration again demonstrates its superiority by achieving the highest mIoU of \textcolor{red}{53.6\%}. Our model is not only the most accurate but also highly parameter-efficient. With only \textcolor{blue}{12.58M} parameters, it outperforms much larger models like FactSeg and FarSeg by over 3.5 percentage points in mIoU. Compared to the lightweight UnetFormer-ResNet18 baseline, our model provides a significant +1.2\% mIoU gain for a marginal increase in model size (+0.85M parameters), indicating superior parameter utilization. LWGANet-L2's strong performance on diverse classes like \textit{Barren} and \textit{Forest} further validates its effectiveness for complex land-cover mapping tasks.

\begin{table*}[h]	\centering \scriptsize
	\renewcommand{\arraystretch}{1.2}
	\setlength{\tabcolsep}{2mm}{
		\begin{tabular}{c c c c|ccccccc}\toprule
			\multirow{2}{*}{Decoder} & \multirow{2}{*}{Backbone} & \multirow{2}{*}{\makecell[c]{Params. \\ (M) $\downarrow$}} & \multirow{2}{*}{\makecell[c]{mIoU \\ (\%) $\uparrow$}} & \multicolumn{7}{c}{IoU per category (\%) $\uparrow$} \\ \cline{5-11}
			& & & & Background & Building & Road & Water & Barren & Forest & Agriculture \\ \midrule
			FactSeg & ResNet50 & 33.44 & 50.0 & 42.51 & 54.62 & 55.88 & 77.96 & 16.51 & 44.72 & 57.81 \\
			FarSeg & ResNet50 & 31.37 & 50.1 & 43.15 & 55.41 & 55.91 & 78.88 & 16.51 & 43.94 & 56.96 \\
			UnetFormer & ResNet18 & \textcolor{red}{11.73} & \textcolor{blue}{52.4} & 44.7 & 58.8 & 54.9 & 79.6 & \textcolor{blue}{20.1} & \textcolor{blue}{46.0} & 62.5 \\
			RSSFormer & RSS-B & 30.82 & \textcolor{blue}{52.4} & \textcolor{blue}{52.38} & \textcolor{red}{60.71} & 55.21 & 76.29 & 18.73 & 45.39 & 58.33 \\
			LoveNAS & ResNet50 & 30.49 & 52.3 & 45.39 & 58.86 & \textcolor{red}{59.45} & 79.39 & 13.76 & 43.94 & \textcolor{red}{65.64} \\
			\rowcolor{gray!30} UnetFormer & \textbf{LWGANet-L2} & \textcolor{blue}{12.58} & \textcolor{red}{53.6} & \textcolor{red}{46.76} & \textcolor{blue}{59.56} & \textcolor{blue}{56.73} & \textcolor{blue}{79.57} & \textcolor{red}{23.58} & \textcolor{red}{46.28} & \textcolor{blue}{62.42} \\ \bottomrule
	\end{tabular}}
	\caption{Semantic segmentation results on the LoveDA test set, comparing with state-of-the-art models.} \label{seg-LoveDA}
\end{table*}

\section{Qualitative Experimental Results}

\subsection{Object Detection Visualization}
\begin{figure*}[h]
	\centering
	\includegraphics[width=1\linewidth]{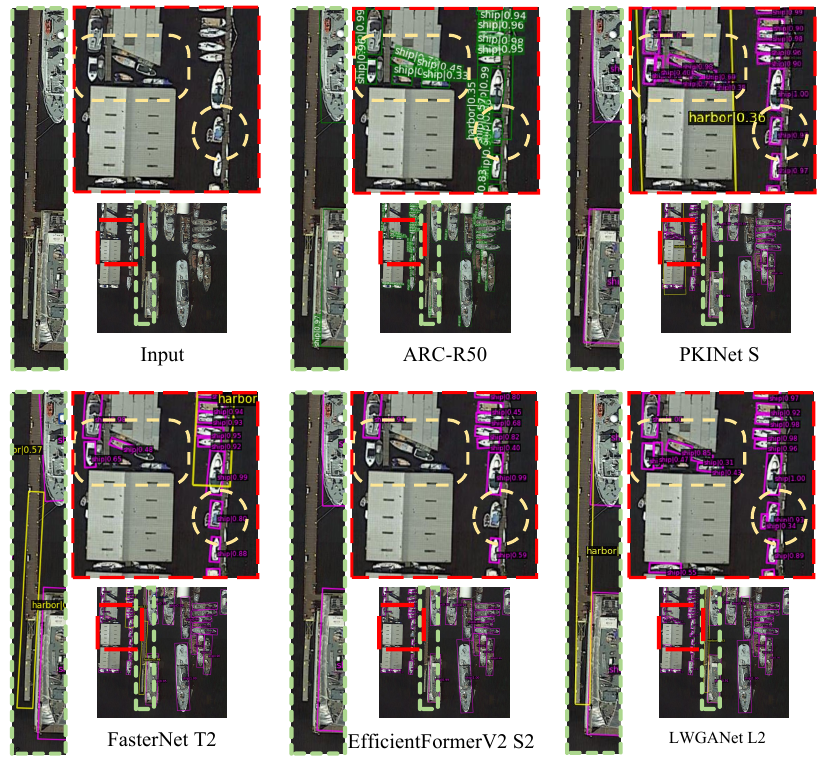}
	\caption{Qualitative comparison on the DOTA 1.0 test set. All backbones are integrated with the Oriented R-CNN detector. LWGANet demonstrates superior performance in detecting multi-scale objects, correctly identifying both large harbors and small ships, whereas other methods exhibit missed detections or inaccurate localizations.} \label{visual_obb}
\end{figure*}

Figure \ref{visual_obb} provides qualitative results for oriented object detection on the DOTA 1.0 test set. We compare LWGANet-L2 against other prominent backbones, including ARC-R50~\cite{ARC}, PKINet~\cite{cai2024pkinet}, FasterNet~\cite{chen2023run}, and EfficientFormerV2~\cite{Eff_Formerv2}. The visualizations clearly demonstrate LWGANet's superior capability in handling objects with significant scale variations—a common challenge in aerial imagery. Our model accurately detects and delineates objects of vastly different sizes within the same scene, such as large harbors, medium-sized vessels, and small vehicles. In contrast, other lightweight backbones like FasterNet and EfficientFormerV2 exhibit robustness issues, such as missed detections for small vehicles and coarse, inaccurate bounding boxes for large vessels. In contrast, LWGANet's effective multi-scale feature fusion allows it to maintain high sensitivity to small targets while preserving precise localization for large objects, as evidenced by the tight and accurate bounding boxes. This visual evidence corroborates our quantitative findings and highlights the effectiveness of LWGANet's multi-scale feature representation.

\clearpage
\subsection{Semantic Segmentation Visualization}
\begin{figure*}[h]
	\centering
	\includegraphics[width=1\linewidth]{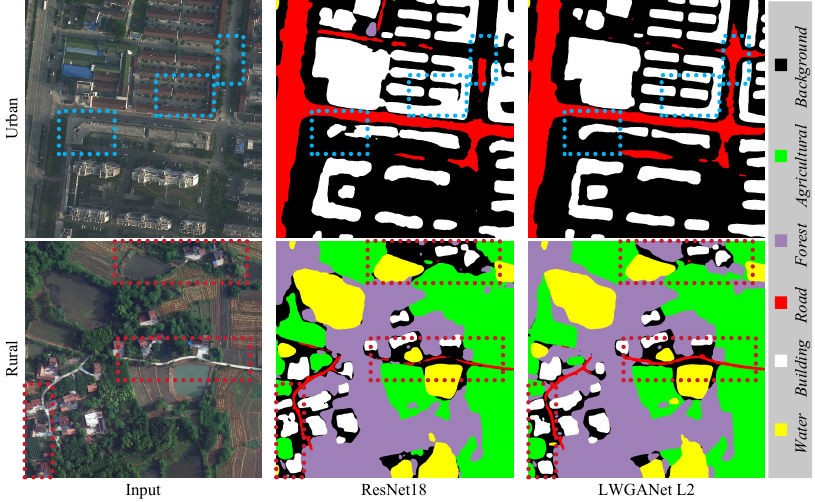}
	\caption{Qualitative results on the LoveDA test set, using UnetFormer as the segmentation head. LWGANet produces cleaner segmentation maps with more precise boundaries for roads and buildings, demonstrating its superior capability in capturing fine-grained details compared to other backbones.} \label{LoveDA_visual}
\end{figure*}

The qualitative results for semantic segmentation on the LoveDA test set are shown in Figure \ref{LoveDA_visual}. The visualizations highlight the distinct advantage conferred by the LWGA module. LWGANet produces segmentation maps with remarkably sharp and accurate boundaries, especially for fine-grained structures like roads and buildings. Compared to other methods, our model generates more coherent and detailed predictions, minimizing noise and correctly delineating complex geometries. This visual superiority underscores the efficacy of our multi-scale attention mechanism in capturing the intricate spatial details crucial for high-fidelity semantic segmentation in RS imagery.

\clearpage
\section{Ablation Study Results}
To validate the contribution of each component within the (\texttt{LWGA module}), we conducted a comprehensive ablation study, the results of which are presented in Table \ref{table_RSVT_abl}. The study evaluates the impact of the GPA, RLA, SMA, and SGA modules across four downstream tasks: classification, detection, segmentation, and change detection.

Our baseline model, shown in the first row, utilizes only the RLA module for feature extraction. Subsequent rows show the performance as we systematically remove one attention component at a time from the full LWGANet-L2 architecture (final row). To isolate the architectural contributions and ensure experimental efficiency, all models in this study were trained from scratch without ImageNet pre-training.

The results consistently demonstrate the positive contribution of each module. For instance, removing any single component generally leads to a performance drop across most tasks. The full LWGANet architecture, which combines all four attention mechanisms, achieves the highest performance in classification (96.17\% accuracy), detection (71.59\% mAP), and change detection (83.95\% IoU). Interestingly, while removing the SGA module slightly improved segmentation performance in this specific setting, its inclusion proved crucial for tasks demanding broader contextual reasoning, such as classification and detection. This study validates that our multi-scale attention design, integrating global, regional, local, and channel-wise information, is highly effective for a diverse range of RS visual tasks.

\begin{table*}[h] \centering	\scriptsize
	\renewcommand{\arraystretch}{1.15}
	\setlength{\tabcolsep}{1.6mm}{
		\begin{tabular}{cccc|cccc|cc|cc|cc} \toprule
			\multirow{2}{*}{GPA} & \multirow{2}{*}{RLA} & \multirow{2}{*}{SMA} & \multirow{2}{*}{SGA} & \multicolumn{4}{c|}{Classification (NWPU val)} & \multicolumn{2}{c|}{Detection (DOTA 1.0 val)} & \multicolumn{2}{c|}{Segmentation (UAVid test)} & \multicolumn{2}{c}{Change Detection (LEVIR-CD test)} \\ \cline{5-14}
			& & & & Params. & FLOPs & Acc. (\%) $\uparrow$ & FPS $\uparrow$ & Params. $\downarrow$ & mAP (\%) $\uparrow$ & Params. $\downarrow$ & mIoU (\%) $\uparrow$ & Params. $\downarrow$ & IoU (\%) $\uparrow$ \\ \midrule
			& \ding{51} & & & 27.69 & 4.35 & 95.35 & 3819 & 43.9 & 69.64 & 27.2 & 62.03 & 28.96 & 83.66 \\ \hline
			\ding{51} & \ding{51} & \ding{51} & & 12.54 & 1.84 & 95.83 & \textcolor{red}{3983} & 28.6 & 70.56 & 12.1 & \textcolor{red}{62.16} & 13.80 & 83.23 \\
			\ding{51} & \ding{51} & & \ding{51} & 12.90 & 1.87 & 95.51 & \textcolor{blue}{3942} & 29.1 & 71.10 & 12.4 & 61.82 & 14.17 & 83.73 \\
			\ding{51} & & \ding{51} & \ding{51} & \textcolor{red}{11.95} & \textcolor{red}{1.70} & 95.86 & 3490 & \textcolor{red}{28.2} & 70.09 & \textcolor{red}{11.5} & 61.71 & \textcolor{red}{13.21} & 83.50 \\
			& \ding{51} & \ding{51} & \ding{51} & \textcolor{blue}{12.06} & \textcolor{blue}{1.71} & \textcolor{blue}{95.98} & 3638 & \textcolor{blue}{28.3} & \textcolor{blue}{71.16} & \textcolor{blue}{11.6} & 61.90 & \textcolor{blue}{13.33} & \textcolor{blue}{83.74} \\
			\rowcolor{gray!30} \ding{51} & \ding{51} & \ding{51} & \ding{51} & 13.01 & 1.87 & \textcolor{red}{96.17} & 3308 & 29.2 & \textcolor{red}{71.59} & 12.6 & \textcolor{blue}{62.05} & 14.30 & \textcolor{red}{83.95} \\ \bottomrule
	\end{tabular}}
	\caption{Ablation study of attention components in the LWGA Block. The first row represents a baseline using only the RLA module. The final row is the full LWGANet-L2 model. Decoders used are Oriented R-CNN (Detection), UnetFormer (Segmentation), and A2Net (Change Detection).} \label{table_RSVT_abl}
\end{table*}

\section{Further Design Rationale and Discussion} \label{sec:appendix_discussion}

This appendix provides additional details and discussions to complement the main paper. We aim to offer deeper insights into the design choices of LWGANet, elaborate on the relationship between our work and existing methods, and discuss the broader applicability of our proposed principles. 
\subsection{Relationship to Prior Works} \label{sec:appendix_novelty}

\paragraph{LWGA: From Homogeneous Grouping to Heterogeneous Multi-Scale Representation.}
Channel grouping is a well-established strategy for building efficient neural networks, as exemplified by the grouped convolutions in ShuffleNet~\cite{zhang2018shufflenet}, depthwise separable convolutions in MobileNets~\cite{mobilenetv2}, and multi-head self-attention in Vision Transformers~\cite{vit}. A common characteristic of these methods is their reliance on \textbf{homogeneous operations}, where each channel group is processed by identical, replicated computational units. While effective for general-purpose vision tasks, this paradigm proves suboptimal for RS imagery, which is characterized by extreme variations in object scale and morphology. In such contexts, a uniform feature extraction strategy leads to a specific form of representational inefficiency: channels become overly specialized to certain scales, resulting in redundancy and a diminished capacity to model the full spectrum of visual concepts present in RS data.

Our LWGA module introduces a departure from this homogeneous paradigm. It implements a \textbf{heterogeneous, multi-scale architecture} within a single block. Instead of merely parallelizing identical operations, LWGA partitions the feature channels and routes each subset through a distinct, computationally specialized pathway, each engineered to capture features at a specific scale---from point-wise details to global context. This design moves beyond simple ensembling, representing a detailed approach to \textbf{decouple the multi-scale representation learning problem into specialized, non-competing sub-tasks}. By assigning specialized operators to different channel subspaces, LWGA mitigates inter-scale interference and promotes a more comprehensive feature representation. The state-of-the-art results across diverse RS tasks validate this design's efficacy. More broadly, our work advocates for heterogeneous designs as a powerful tool for building efficient models for domains with high intra-class and inter-scale variance, offering a compelling alternative to conventional, homogeneous architectures.

\paragraph{TGFI: A Lightweight Mechanism for Sparse Global Interaction.}
The quadratic complexity of self-attention has spurred extensive research into sparse attention mechanisms. Our TGFI module aligns with this research direction but is differentiated by its design principles of simplicity and non-parametric efficiency. While many existing token sparsification methods employ learnable routing modules or computationally intensive clustering algorithms—introducing their own parametric or computational overhead—TGFI adopts a \textbf{simple, non-parametric sampling strategy based on feature activation magnitudes, making it computationally lean}. This approach is particularly well-suited for RS data, where salient foreground objects are often sparsely distributed across vast, low-information backgrounds. Critically, TGFI is not designed as a standalone attention replacement but as an integral component of the LWGA block. It functions as an efficient mechanism to enable a full-fledged global attention pathway on a reduced set of high-value tokens, ensuring that LWGANet can model long-range dependencies without prohibitive computational cost.

\subsection{Discussion on Potential Broader Impacts} \label{sec:appendix_generality}
The core challenge addressed in this work—efficiently modeling data with high spatial redundancy and channel redundancy—is not unique to RS. We note that similar data characteristics are prevalent in other important domains. For instance, digital pathology with its gigapixel-scale images and high-resolution document analysis present analogous challenges of vast, uninformative backgrounds interspersed with critical, multi-scale features.

The design of LWGANet, particularly the idea of decoupling feature representation into specialized pathways, might offer a valuable perspective for researchers in these fields. While a direct application would require domain-specific adaptations and validation, we believe our work could serve as a conceptual starting point. We hope that our findings encourage the exploration of similar function-specialized, heterogeneous architectures for efficient visual analysis beyond the scope of RS.

\bibliography{lwganet}

\end{document}